\PassOptionsToPackage{capitalize}{cleveref}
\documentclass[10pt,twocolumn,letterpaper]{article}

\usepackage{cvpr}

\usepackage{graphicx}
\usepackage{amsmath,amssymb}
\usepackage{booktabs}
\usepackage{subcaption}
\usepackage[table]{xcolor}
\usepackage{tabularx}
\usepackage{siunitx}
\usepackage{algorithm}
\usepackage[noend]{algpseudocode}
\definecolor{cvprblue}{rgb}{0.21,0.49,0.74}
\usepackage[pagebackref,breaklinks,colorlinks,allcolors=cvprblue]{hyperref}
\usepackage{cleveref}
\crefname{figure}{Fig.}{Figs.}
\Crefname{figure}{Fig.}{Figs.}
\crefname{table}{Tab.}{Tabs.}
\Crefname{table}{Tab.}{Tabs.}
\crefname{section}{Sec.}{Secs.}
\Crefname{subsection}{Sec.}{Secs.}


\begin{document}

\title{CritiFusion: Semantic Critique and Spectral Alignment for Faithful Text-to-Image Generation}

\author{
ZhenQi Chen$^{1}$ \quad
TsaiChing Ni$^{1}$ \quad
YuanFu Yang$^{1}$\\
$^{1}$National Yang Ming Chiao Tung University\\
{\tt\small \{alex.ii13, nina.ii13, yfyangd\}@nycu.edu.tw}
}

\twocolumn[{
\maketitle
\begin{center}
    \vspace{-2mm}
    \includegraphics[width=0.95\textwidth]{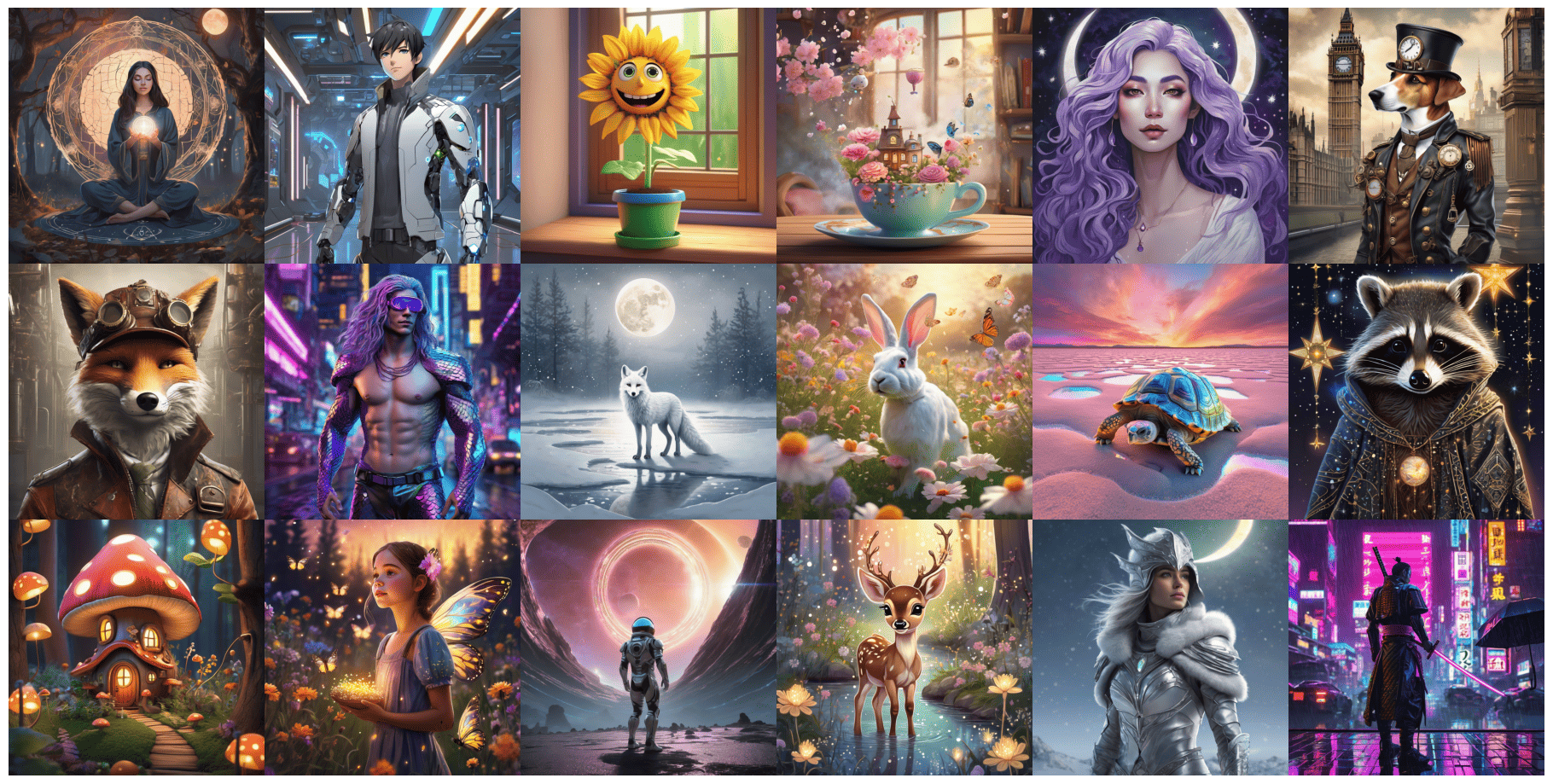}
    \vspace{-2mm}
    \captionof{figure}{
        \textbf{Diverse high-fidelity generations by our CritiFusion model.} 
        Our method synthesizes diverse artistic and photorealistic scenes with improved prompt alignment and high visual fidelity.
    }
    \label{fig:teaser}
    \vspace{2mm}
\end{center}
}]
\begin{abstract}
\noindent
Recent text-to-image diffusion models have achieved remarkable visual fidelity but often struggle with semantic alignment to complex prompts. We introduce CritiFusion, a novel inference-time framework that integrates a multimodal semantic critique mechanism with frequency-domain refinement to improve text-to-image consistency and detail. The proposed CritiCore module leverages a vision-language model and multiple large language models to enrich the prompt context and produce high-level semantic feedback, guiding the diffusion process to better align generated content with the prompt’s intent. Additionally, SpecFusion merges intermediate generation states in the spectral domain, injecting coarse structural information while preserving high-frequency details. No additional model training is required. CritiFusion serves as a plug-in refinement stage compatible with existing diffusion backbones. Experiments on standard benchmarks show that our method notably improves human-aligned metrics of text-to-image correspondence and visual quality. CritiFusion consistently boosts performance on human preference scores and aesthetic evaluations, achieving results on par with state-of-the-art reward optimization approaches. Qualitative results further demonstrate superior detail, realism, and prompt fidelity, indicating the effectiveness of our semantic critique and spectral alignment strategy.
\end{abstract}

\section{Introduction}
\label{sec:intro}

\noindent
Text-to-image (T2I) diffusion models can synthesize diverse, high-quality images from natural-language prompts~\cite{ldm, dalle, freedom, imagen, parti, sdxl}. Yet semantic alignment—faithfully realizing the textual intent in generated images—remains challenging, especially under compositional prompts with multiple entities, attributes, and relations. Typical failures fall into three measurable types, namely (i) attribute binding errors (e.g., color-object mismatch), (ii) relational or positional errors (e.g., left-of or behind), and (iii) fine-detail omissions. These errors persist because large-scale training is weakly supervised, and the inference objective balances realism against textual faithfulness in ways that are not directly optimized.

Recent efforts address alignment along two directions. One line pursues training-based preference alignment using human feedback or learned reward models~\cite{ rewardbp, rl-diffusion, aligninghf}, including direct preference optimization within the diffusion process~\cite{diffusiondpo, spo}. The other line explores training-free control at inference, such as dynamic guidance and prompt manipulation~\cite{dymo, prompts-opt, recaption}. Training-based methods often improve human-preference or aesthetic scores~\cite{hpsv2, imagereward, laion} but require additional optimization and may overfit to specific reward models or data biases. Training-free methods avoid retraining but typically rely on scalar rewards that are unreliable at early denoising steps when the image is still highly ambiguous.

We propose CritiFusion, a training-free inference-time framework that couples high-level semantic critique with frequency-domain fusion to improve semantic alignment without modifying backbone weights. The core idea is to inject multimodal feedback into the generation loop. A CritiCore module aggregates judgments from a pre-trained Vision-Language Model (VLM) and Multiple Large Language Models (Multi-LLM) agents to identify omissions and ambiguities from the prompt-image pair and to produce a refined textual conditioning. In practice, we map the critique to a residual update on the text-conditioning pathway (cross-attention conditioning), yielding an enriched embedding that emphasizes entities, attributes, and relations most at risk of failure. Complementing this semantic pathway, SpecFusion performs frequency-domain fusion between the initial draft and the refined generation, where a smooth low-frequency gate preserves global layout and lighting while high-frequency components inject corrected local details. This design stabilizes composition and enables targeted edits, and smooth spectral masks mitigate halo and ringing artifacts.

Our framework is plug-and-play and applies to common T2I backbones. Figure~\ref{fig:teaser} illustrates representative before and after examples under compositional prompts. We evaluate on public benchmarks with HPSv2, ImageReward, and aesthetic metrics~\cite{hpsv2, imagereward, laion}, and we provide qualitative analyses and ablations to ensure that improvements are not artifacts of automated scorers. In experiments, we show that CritiFusion is comparable to or better than strong preference-aligned baselines on most metrics while preserving aesthetic quality and requiring no retraining.

This work makes three contributions:
\begin{itemize}
\item \textbf{CritiCore.} A multi-agent semantic critique that fuses VLM and LLM feedback into a residual update of text conditioning, targeting measurable failure types (attribute binding, relations, fine details) under compositional prompts.
\item \textbf{SpecFusion.} A frequency-domain fusion mechanism that preserves low-frequency structure for layout and illumination and injects high-frequency corrections, which enables stable global composition with localized improvements.
\item \textbf{Training-free, backbone-agnostic validation.} A plug-in refinement stage that requires no weight updates. We report consistent gains across multiple backbones and datasets, and we document compute-quality trade-offs and limitations.
\end{itemize}

\begin{figure*}[t]
\centering
\includegraphics[width=\textwidth]{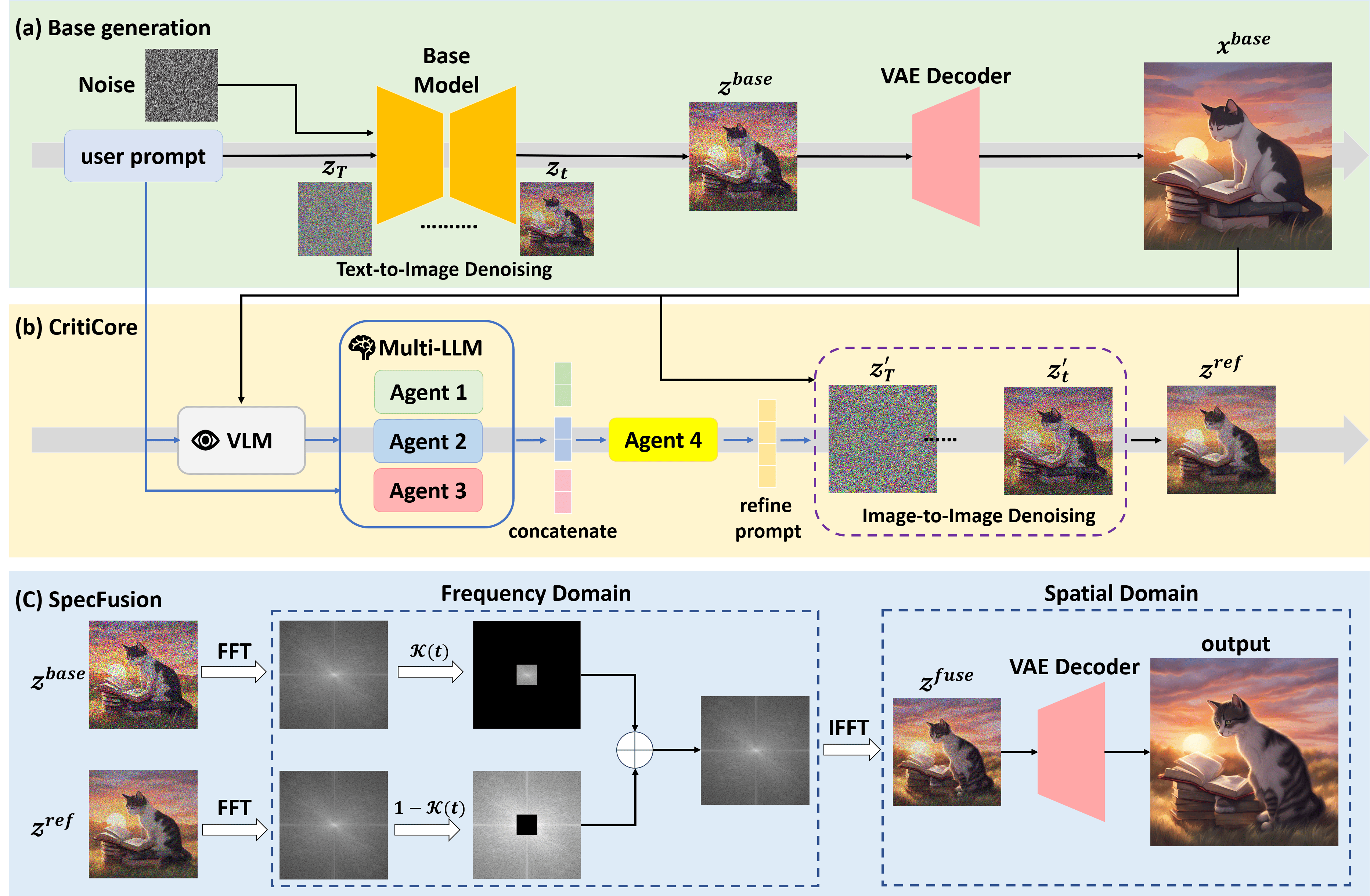}
\caption{\textbf{Overview of our framework.}
 Our pipeline has two modules. (1) CritiCore fuses VLM captions with Multi-LLM feedback to produce a refined prompt embedding $\tilde{c}$. A diffusion backbone samples a base latent $\mathbf{z}^{\text{base}}$ (decoded as $\mathbf{x}^{\text{base}}$) and, guided by $\tilde{c}$, partially re-denoises it to $\mathbf{z}^{\text{ref}}$. (2) SpecFusion performs frequency gating: the high-frequency spectrum of $\mathbf{z}^{\text{ref}}$ is combined with the low-frequency spectrum of $\mathbf{z}^{\text{base}}$ to yield $\tilde{\mathbf{z}}$, which is decoded to the final image $\hat{\mathbf{x}}$. This two-stage refinement corrects semantic misalignment, enriches detail, and preserves global coherence.}
\label{fig:arch}
\end{figure*}

\section{Related Work}
\label{sec:related}

\noindent\textbf{Text-to-Image Diffusion Models.}
Diffusion models synthesize high-fidelity, diverse images by reversing a forward noising process~\cite{ddpm}. Latent diffusion scales this paradigm by operating in a learned latent space with markedly lower computational cost~\cite{ldm}. Early text-conditional systems such as DALL-E~\cite{dalle} and GLIDE~\cite{glide} established strong baselines, followed by Imagen, Parti, SDXL, and eDiff which advanced photorealism and resolution~\cite{imagen,parti,sdxl,ediffi}. Persistent errors in fine-grained and compositional semantics motivate inference-time strategies that enhance prompt adherence without retraining. Our approach injects structured guidance during sampling and improves both alignment and perceptual quality.

\noindent\textbf{Alignment via Human Feedback.}
Preference alignment for diffusion includes differentiable reward fine-tuning and RL-style objectives optimized against learned or human-derived signals~\cite{draft,rewardbp,rl-diffusion,instructgpt}. Direct Preference Optimization adapts LLM preference learning to diffusion (Diffusion-DPO)~\cite{diffusiondpo}, and utility or KTO-style objectives also report gains~\cite{d3po}. Benchmarks increasingly rely on large-scale preference predictors such as HPS and HPSv2~\cite{hps,hpsv2}. Although training-based methods can be effective, they require additional optimization and may bias models toward the reward. Training-free approaches that reweight or schedule guidance at inference (e.g., DyMO)~\cite{dymo} show that part of the alignment gap can be closed without modifying weights. We extend this line by coupling inference-time control with multimodal critique that targets semantic errors explicitly.

\noindent\textbf{Prompt Enhancement and Multimodal Feedback.}
Prompt-level interventions adapt or enrich textual conditioning to better match model priors~\cite{prompts-opt,recaption}. In addition, VLM offer image-grounded feedback. CLIP provides transferable joint representations~\cite{clip}, and coordination frameworks such as HuggingGPT and Visual ChatGPT demonstrate LLM-VLM collaboration for planning, critique, and revision~\cite{hugginggpt,visual-chatgpt,gpt4}. Building on these insights, CritiCore employs an LLM committee for semantic enrichment and a VLM for image-grounded critique. The revisions are re-encoded into the diffusion backbone at inference, which strengthens attributes and relations while preserving the generator’s aesthetics.

\section{Method}
\label{sec:method}
\vspace{-2pt}
\noindent
Our CritiFusion augments a pretrained latent diffusion model with two inference-time modules as shown in Figure~\ref{fig:arch}. CritiCore refines textual conditioning via multimodal feedback, and SpecFusion enforces frequency-domain consistency on the latent representation. Both operate at inference with frozen backbones. Algorithm~\ref{alg:critifusion-full} outlines one refinement round in which a base pass produces a provisional image, a short img2img corrective pass is governed by CADR, and SpecFusion merges spectral content to recover stable global appearance. The cost is one base diffusion pass and one corrective pass. FFT overhead is negligible. No architectural changes or retraining are required. We fix the random seed across passes in ablations. In practice, the method integrates through a lightweight sampler wrapper and preserves near-baseline throughput on a single GPU.

\subsection{Preliminaries}
\noindent
Let $\mathbf{z}_T\!\sim\!\mathcal{N}(\mathbf{0},\mathbf{I})$ and $\varepsilon_\theta(\mathbf{z}_t,t\mid c)$ be the U\!-Net noise predictor under prompt $c$. With a variance-preserving schedule $\{\beta_t\}_{t=1}^T$, the reverse step~\cite{ddpm} is
\begin{equation}
\mathbf{z}_{t-1}
=\frac{1}{\sqrt{1-\beta_t}}\Bigl(\mathbf{z}_t-\beta_t\,\varepsilon_\theta(\mathbf{z}_t,t\mid c)\Bigr)
+\sigma_t\,\boldsymbol\omega,
\label{eq:ddpm}
\end{equation}
where $\boldsymbol\omega\!\sim\!\mathcal{N}(\mathbf{0},\mathbf{I})$ and $\sigma_t^2$ is the posterior variance. The analysis is sampler agnostic. For ablations the random seed is fixed. We keep the base and corrective passes deterministic to isolate the effect of each module.

\vspace{4pt}
\noindent
A first pass with $c$ yields the base latent $\mathbf{z}^{\text{base}}$ and the image $\mathbf{x}^{\text{base}}=\mathcal{D}(\mathbf{z}^{\text{base}})$. CritiCore revises the conditioning and runs a short img2img pass from $\mathbf{z}^{\text{base}}$ to produce the refined latent $\mathbf{z}^{\text{ref}}$. SpecFusion then merges $\mathbf{z}^{\text{ref}}$ and $\mathbf{z}^{\text{base}}$ in the frequency-domain to obtain $\tilde{\mathbf{z}}$, which decodes to the final image.

\subsection{CritiCore for Multi-LLM Semantic Refinement}

\paragraph{VLM grounded critique.}
Given $\mathbf{x}^{\text{base}}$, a VLM produces concise, evidence-grounded hints $r_{\mathrm{cc}}$ for prompt $c$ e.g., missing entities, malformed attributes, and spatial misplacements. These hints guide clause decomposition and per-clause scoring, which stabilizes the corrective pass. We keep top $k$ hints with $k\!=\!5$ and calibrate scores to a common range. This keeps edits localized and reduces prompt drift.

\paragraph{Prompt decomposition.}
A committee of LLMs $\{\mathcal{T}_j\}_{j=1}^M$ maps $(c,r_{\mathrm{cc}})$ to grounded visual clauses $\mathcal{C}=\{\tau_i\}_{i=1}^m$ covering entities, attributes, and relations. An aggregator $\mathcal{T}_{\text{agg}}$ deduplicates, resolves conflicts, and orders by salience to yield $\tilde c_0$. The VLM scores each clause $s_i=f_{\text{vlm}}(\tau_i,\mathbf{x}^{\text{base}})$. The lowest scoring clauses \emph{top $k$} are selected as edits, and the Aggregator outputs the final conditioning $\tilde c$. A token budget $K$ is enforced by pruning low salience paraphrases and keeping minimal phrasing to preserve cross attention capacity.

\paragraph{Partial Re-denoising with CADR}
\noindent
We correct from $\mathbf{z}^{\text{base}}$ instead of restarting from noise. The Controlled Adaptive Denoising Rate maps the mean VLM score $s=\frac{1}{m}\sum_i s_i$ to the inference hyperparameters
\begin{equation}
(\lambda,\ g,\ T',\ \rho)\;=\;\Phi(s)\;=\;A + (1-s)\,B,
\label{eq:cadr}
\end{equation}
where $\lambda$ is the img2img strength, $g$ is the CFG scale, $T'$ is the corrective step count, and $\rho$ is the SpecFusion mask parameter. Lower $s$ increases edit strength with larger $\lambda$, $g$, and $T'$, and it tightens low-frequency preservation through $\rho$. We use an affine schedule with $\lambda\!\in\![0.12,0.30]$, $g\!\in\![3.6,5.0]$, $T'\!\in\![16,30]$, and $\rho\!\in\![0.60,0.85]$. Let $\mathcal{R}_{T',g}$ denote $T'$ solver steps under guidance $g$ and conditioning $\tilde c$. Then
\begin{equation}
\mathbf{z}^{\text{ref}}
=
\mathcal{R}_{T',g}\!\bigl(\mathbf{z}^{\text{base}} \,\big|\, \tilde c,\ \lambda \bigr).
\label{eq:refine}
\end{equation}
Modulating $\lambda$ preserves geometry, lighting, and composition while fixing local semantic errors. We clamp $\Phi(s)$ to the above ranges and enforce monotonicity. The img2img start step $t_0$ is inferred from $\lambda$ through the sampler strength to step mapping to maintain consistent noise levels. During batched sampling, $s$ can be lightly smoothed across repeats of the same prompt. If $s$ is greater than $0.9$, we set $T'\!=\!0$ to skip refinement.

\subsection{SpecFusion for Frequency-Domain Consistency}
\noindent
SpecFusion separates low and high frequency content to mitigate global shifts. With $\mathcal{F}$ the 2D FFT, let $\mathbf{Z}_{\rm lo}=\mathcal{F}(\mathbf{z}^{\text{base}})$ and $\mathbf{Z}_{\rm hi}=\mathcal{F}(\mathbf{z}^{\text{ref}})$. We fuse with a confidence controlled mask
\begin{equation}
\mathbf{Z}_{\mathrm{fuse}}
=\underbrace{K(\rho)\odot\mathbf{Z}_{\rm lo}}_{\text{global layout}}
+\underbrace{\bigl(1-K(\rho)\bigr)\odot\mathbf{Z}_{\rm hi}}_{\text{fine details}},
\label{eq:fm}
\end{equation}
where $K(\rho)$ is a low pass rectangular mask whose passband increases with $\rho$, and $\odot$ denotes element-wise multiplication. The final latent is
\begin{equation}
\tilde{\mathbf{z}}=\mathcal{F}^{-1}(\mathbf{Z}_{\mathrm{fuse}}).
\label{eq:ifft}
\end{equation}
Low frequencies preserve layout, illumination, and color from $\mathbf{z}^{\text{base}}$, while high frequencies import fixes from $\mathbf{z}^{\text{ref}}$. Linking $K(\rho)$ to $s$ aligns spectral trust with semantic confidence. Compared with spatial residual skips, the frequency selective merge retains detail and reduces ringing near edges at the cost of two forward FFTs, one blend, and one inverse FFT. A cosine-tapered boundary on $K(\rho)$ further reduces Gibbs artifacts. Operations are per-channel using real FFTs. Hermitian symmetry is preserved and spectral energy is clipped to the empirical range of $\mathbf{z}^{\text{base}}$ before decoding.

\begin{algorithm}[t]
\caption{CritiFusion Inference}
\label{alg:critifusion-full}
\small
\begin{algorithmic}[1]
\Require prompt $c$, backbone sampler $\mathrm{BaseSampler}$, token budget $K$
\Ensure final image $\hat{x}$
\State $\mathbf z^{\mathrm{base}} \leftarrow \mathrm{BaseSampler}(c)$
\State $\mathbf x^{\mathrm{base}} \leftarrow \mathcal D(\mathbf z^{\mathrm{base}})$
\State $r_{\mathrm{cc}} \leftarrow \mathrm{VLMRefine}(\mathbf x^{\mathrm{base}},c)$
\State $\mathcal{C} \leftarrow \mathrm{MultiLLMAgents}(c, r_{\mathrm{cc}})$
\State $\tilde c_0 \leftarrow \mathrm{Aggregator}(\mathcal{C})$
\State $s_i \leftarrow f_{\mathrm{vlm}}(\tau_i,\mathbf x^{\mathrm{base}})$ for $i=1\ldots m$
\State $\tilde c \leftarrow \mathrm{MergeTopK}(\tilde c_0, \{s_i\})$
\State $s \leftarrow \frac{1}{m}\sum_{i=1}^m s_i$   
\State $(\lambda, g, T', \rho) \leftarrow \Phi_{\mathrm{CADR}}(s)$
\State $\mathbf z^{\mathrm{ref}} \leftarrow
       \mathrm{Img2ImgRefine}\bigl(\mathbf z^{\mathrm{base}},\tilde c,\lambda,g,T'\bigr)$
\State $\tilde{\mathbf z} \leftarrow
       \mathrm{SpecFusion}\bigl(\mathbf z^{\mathrm{ref}},\mathbf z^{\mathrm{base}},\rho\bigr)$
\State $\hat{x} \leftarrow \mathcal D(\tilde{\mathbf z})$
\State \Return $\hat{x}$
\end{algorithmic}
\end{algorithm}

\vspace{-3pt}
\paragraph{Complexity.} 
The extra cost includes one partial denoising pass($T'\!\le\!30$) and two FFTs per sample. Empirically, the runtime overhead is less than $1.3\times$ compared to the baseline.

\begin{figure*}[t]
  \centering
  \includegraphics[width=\textwidth]{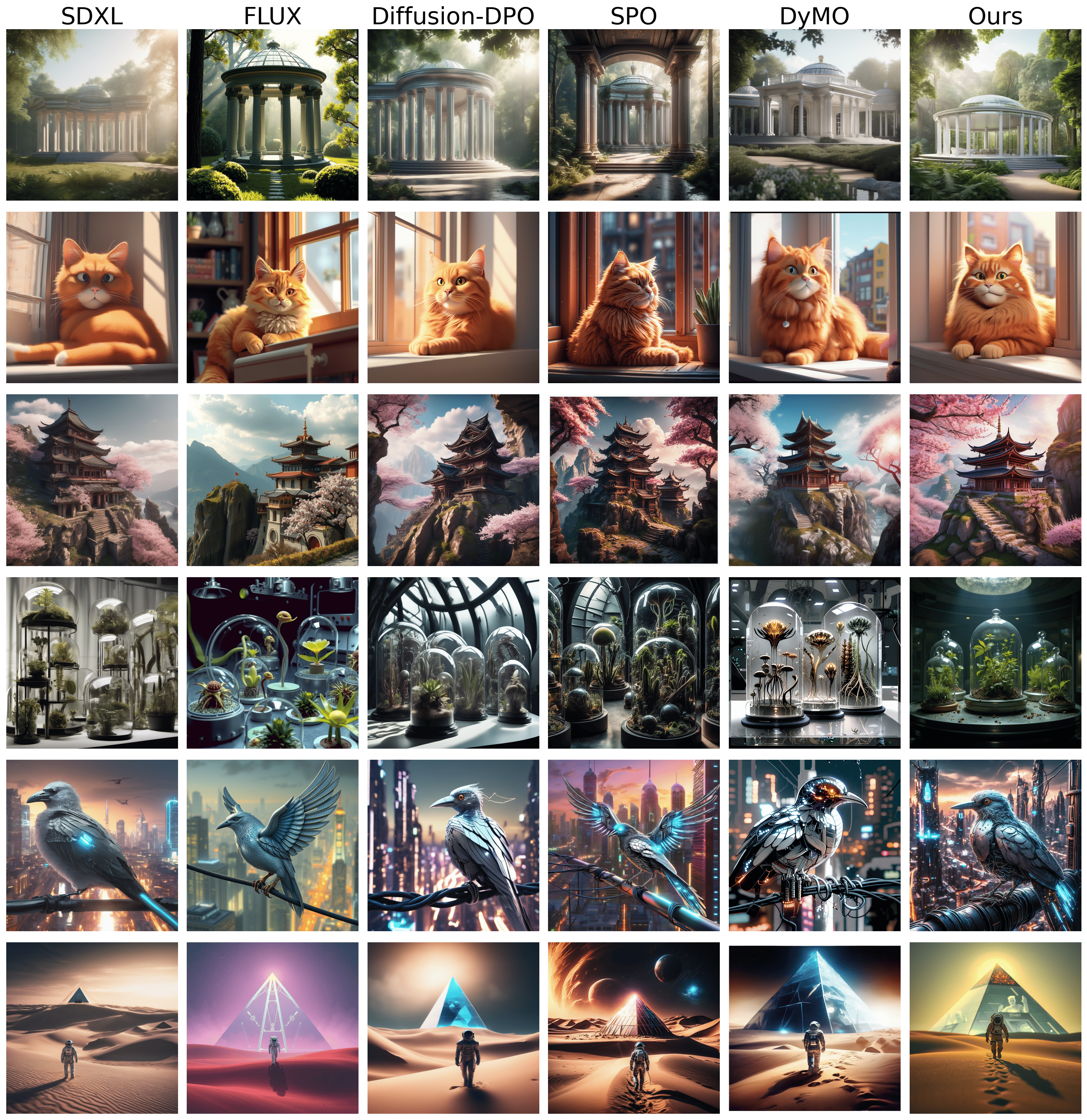}
  \caption{\textbf{Qualitative comparison on various prompts.} We compare images generated by SDXL, FLUX, Diffusion-DPO, SPO, DyMO, and our CritiFusion method across various subjects such as \textit{architecture}, \textit{animals}, \textit{natural scenes}, and \textit{characters}. Each group shares the same prompt for fairness. Our results (far right in each group) show superior realism and semantic alignment: natural color tones, spatial coherence, and faithful prompt adherence. Competing methods often suffer from artifacts or miss subtle details, while CritiFusion maintains balanced composition and high-fidelity rendering. Zoom-in reveals texture and object accuracy, illustrating the benefits of semantic critique and spectral fusion.}
  \label{fig:qualitative}
\end{figure*}

\section{Experiments}
\label{sec:experiments}
\noindent
We evaluate CritiFusion on multiple benchmarks and compare its performance to recent state-of-the-art methods for prompt alignment and image quality. Our experiments demonstrate significant quantitative improvements on learned preference metrics and qualitative enhancements in generated images.

\begin{sloppypar} 
\noindent\paragraph{Implementation details.}
All experiments use SD v1.5~\cite{ldm} and SDXL~\cite{sdxl}. Inference runs for 50 sampling steps on a single NVIDIA A6000, and the random seed is fixed per prompt across all methods to ensure replicability.

\noindent\paragraph{Models.}
We employ a VLM-augmented, multi-agent text stack accessed through the Together Inference API as a remote service, while diffusion backbones run locally. The VLM is \path{meta-llama/Llama-4-Scout-17B-16E-Instruct}. For semantic enrichment we use a mixture of LLMs \path{meta-llama/Llama-3.3-70B-Instruct-Turbo}, \path{Qwen/Qwen2.5-72B-Instruct-Turbo}, \path{Qwen/Qwen2.5-Coder-32B-Instruct}, \path{deepseek-ai/DeepSeek-V3}, and \path{nvidia/NVIDIA-Nemotron-Nano-9B-v2}. A lightweight aggregator (\texttt{AGGREGATOR\_MODEL}, instantiated as \path{Qwen/Qwen2.5-72B-Instruct-Turbo}) fuses agent outputs into the 77-token conditioning. Offloading VLM and LLM inference to Together keeps local VRAM dominated by the diffusion backbone, and SD v1.5 or SDXL pipelines run locally.

\noindent\paragraph{Datasets and metrics.}
Following DyMO~\cite{dymo}, we evaluate on three public prompt suites: \emph{Pick-a-Pic} test (500), \emph{HPSv2} (500), and \emph{PartiPrompts} (1000). We report four higher-is-better automatic metrics aligned with human judgment: \textbf{PickScore}~\cite{pickapic} for overall preference, \textbf{HPSv2}~\cite{hpsv2} for prompt adherence, \textbf{ImageReward}~\cite{imagereward} for text-conditioned quality and relevance, and an \textbf{Aesthetic} predictor~\cite{laion} for text-agnostic visual appeal. This suite separates alignment signals (HPSv2) from general preference (PickScore and ImageReward) and pure aesthetics (Aesthetic).
\end{sloppypar}

\subsection{Quantitative Results}
\label{sec:quant}

\noindent\textbf{SD v1.5 based models.}
As shown in Table~\ref{tab:sd15}, CritiFusion consistently improves SD v1.5 across all evaluation metrics.
Relative to the base model, PickScore rises from 20.73 to 22.02 (+1.29), HPSv2 from 0.234 to 0.264 (+0.030), ImageReward from 0.170 to 0.613 (+0.443), and Aesthetic from 5.34 to 5.86 (+0.52).
These gains are cross metric and substantial, indicating simultaneous improvements in semantic alignment and perceptual fidelity.
CritiFusion also attains an ImageReward close to Diffusion KTO (0.613 vs.\ 0.616), while surpassing it in Aesthetic (5.86 vs.\ 5.70), which shows that improvements do not come at the expense of visual appeal.
In practice, the added cost is modest, since CritiFusion introduces a short img2img corrective pass and lightweight FFTs, so the wall-clock overhead is small under our 50 step DDIM~\cite{ddim}   setting.
Prompts involving spatial relations and multi-attribute bindings benefit the most, consistent with the role of visual critique and spectral stabilization.

Among baselines, DyMO achieves the highest ImageReward (0.717) and strong PickScore and HPSv2 values.
Unlike reward optimized systems, CritiFusion operates entirely at inference time and requires no retraining, which makes its performance gains broadly applicable and low cost.Methods such as PromptOpt and DNO exhibit moderate Aesthetic scores but negative ImageReward values, suggesting challenges in maintaining semantic faithfulness under prompt or reward driven adjustments.
The consistent uplift under identical sampling conditions (50 step DDIM, fixed seeds) highlights the complementary effects of semantic critique and spectral refinement.

\begin{table}[tb]
\centering
\caption{Quantitative comparison on SD \,v1.5-based methods.}
\label{tab:sd15}
\begingroup
\footnotesize
\setlength{\tabcolsep}{3pt}
\renewcommand{\arraystretch}{1.05}
\begin{tabular}{@{}ccccc@{}}
\toprule
\textbf{Methods} & \textbf{PickScore↑} & \textbf{HPSv2↑} & \textbf{ImageReward↑} & \textbf{Aesthetic↑} \\
\midrule
SD v1.5~\cite{ldm}              & 20.73 & 0.234 &  0.170 & 5.34 \\
DNO~\cite{dno}                  & 20.05 & 0.259 & -0.321 & 5.60 \\
PromptOpt~\cite{deckers2023promptopt} & 20.26 & 0.249 & -0.337 & 5.47 \\
FreeDoM~\cite{freedom}          & 21.96 & 0.261 &  0.396 & 5.52 \\
AlignProp~\cite{rewardbp}       & 20.56 & 0.263 &  0.113 & 5.46 \\
Diffusion-DPO~\cite{diffusiondpo} & 20.97 & 0.266 &  0.299 & 5.59 \\
Diffusion-KTO~\cite{diffkto}    & 21.15 & 0.272 &  0.616 & 5.70 \\
SPO~\cite{spo}                  & 21.46 & 0.267 &  0.232 & 5.70 \\
DyMO~\cite{dymo}                & 23.07 & 0.278 &  0.717 & 5.83 \\
\rowcolor[HTML]{E9E9E9}
\textbf{SD v1.5+Ours}           & \textbf{22.02} & \textbf{0.264} & \textbf{0.613} & \textbf{5.86} \\
\bottomrule
\end{tabular}
\endgroup
\end{table}

\begin{table}[tb]
\centering
\caption{Quantitative comparison on SDXL-based methods.}
\label{tab:sdxl}
\begingroup
\footnotesize
\setlength{\tabcolsep}{3pt}
\renewcommand{\arraystretch}{0.98}
\begin{tabular}{@{}ccccc@{}}
\toprule
\textbf{Methods} & \textbf{PickScore↑} & \textbf{HPSv2↑} & \textbf{ImageReward↑} & \textbf{Aesthetic↑} \\
\midrule
SDXL~\cite{sdxl}                 & 21.97 & 0.266 & 0.776 & 5.94 \\
DNO~\cite{dno}                   & 22.14 & 0.273 & 0.905 & 6.04 \\
PromptOpt~\cite{deckers2023promptopt} & 21.98 & 0.271 & 0.867 & 5.88 \\
FreeDoM~\cite{freedom}           & 22.13 & 0.272 & 0.772 & 5.91 \\
DyMO~\cite{dymo}                 & 24.90 & 0.284 & 1.074 & 6.14 \\
\rowcolor[HTML]{E9E9E9}
\textbf{SDXL+Ours}               & \textbf{23.13} & \textbf{0.291} & \textbf{1.076} & \textbf{6.22} \\
Diffusion-DPO~\cite{diffusiondpo} & 22.30 & 0.274 & 0.979 & 5.89 \\
SPO~\cite{spo}                   & 22.81 & 0.278 & 1.082 & 6.32 \\
SPO+DyMO         & 23.85 & 0.282 & 1.166 & 6.28 \\
\rowcolor[HTML]{E9E9E9}
\textbf{SPO+Ours}     & \textbf{24.09} & \textbf{0.304} & \textbf{1.183} & \textbf{6.29} \\
SD v3.5~\cite{ldm}                & 21.93 & 0.273 & 0.970 & 5.78 \\
FLUX~\cite{flux}                 & 22.04 & 0.276 & 1.011 & 6.08 \\
\bottomrule
\end{tabular}
\endgroup
\end{table}

\begin{figure*}[t]
\centering
\begin{subfigure}{0.32\textwidth}
\includegraphics[width=\linewidth]{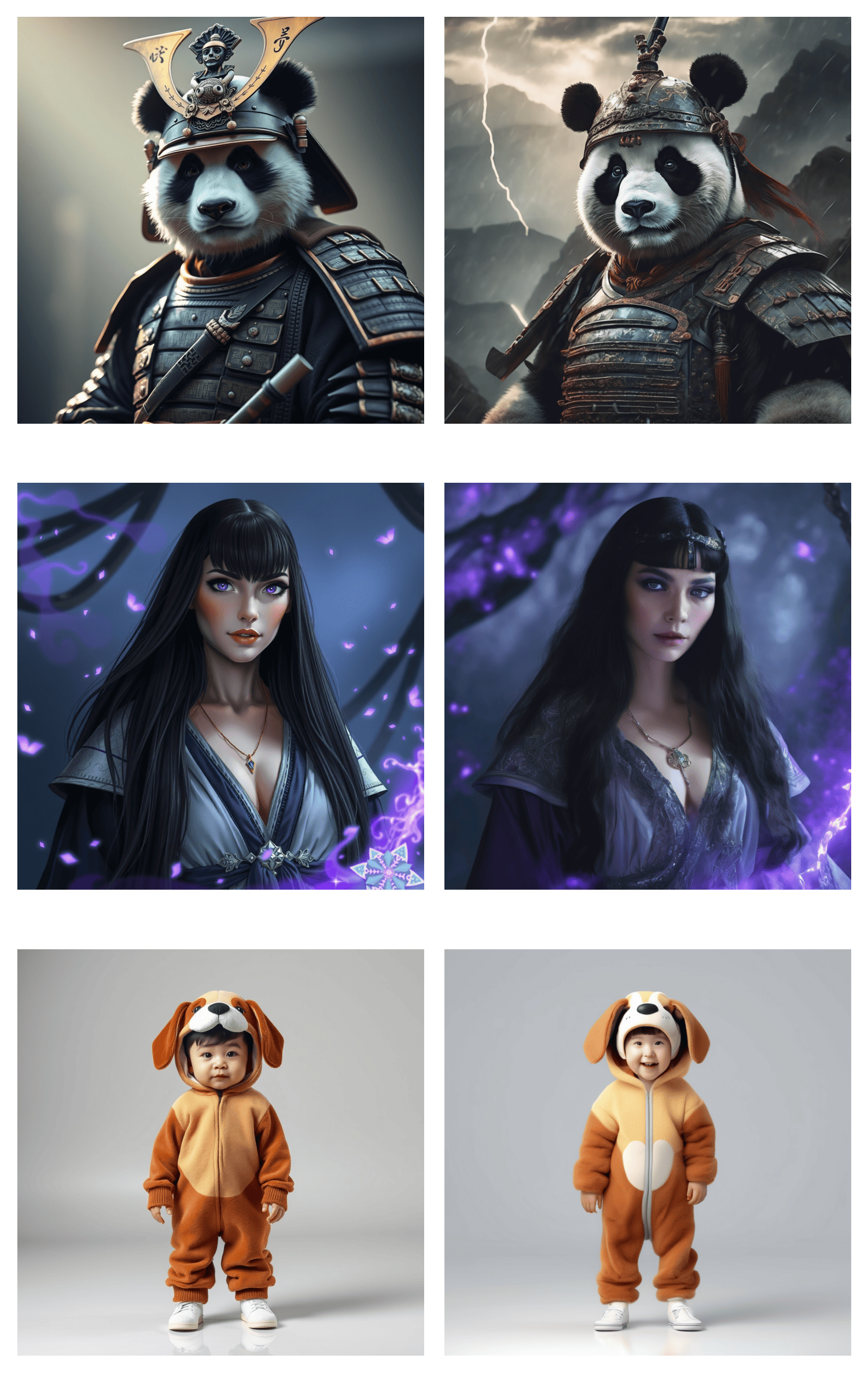}
\caption{\small \textit{FLUX} vs \textit{FLUX + Ours}}
\end{subfigure}
\hfill
\begin{subfigure}{0.32\textwidth}
\includegraphics[width=\linewidth]{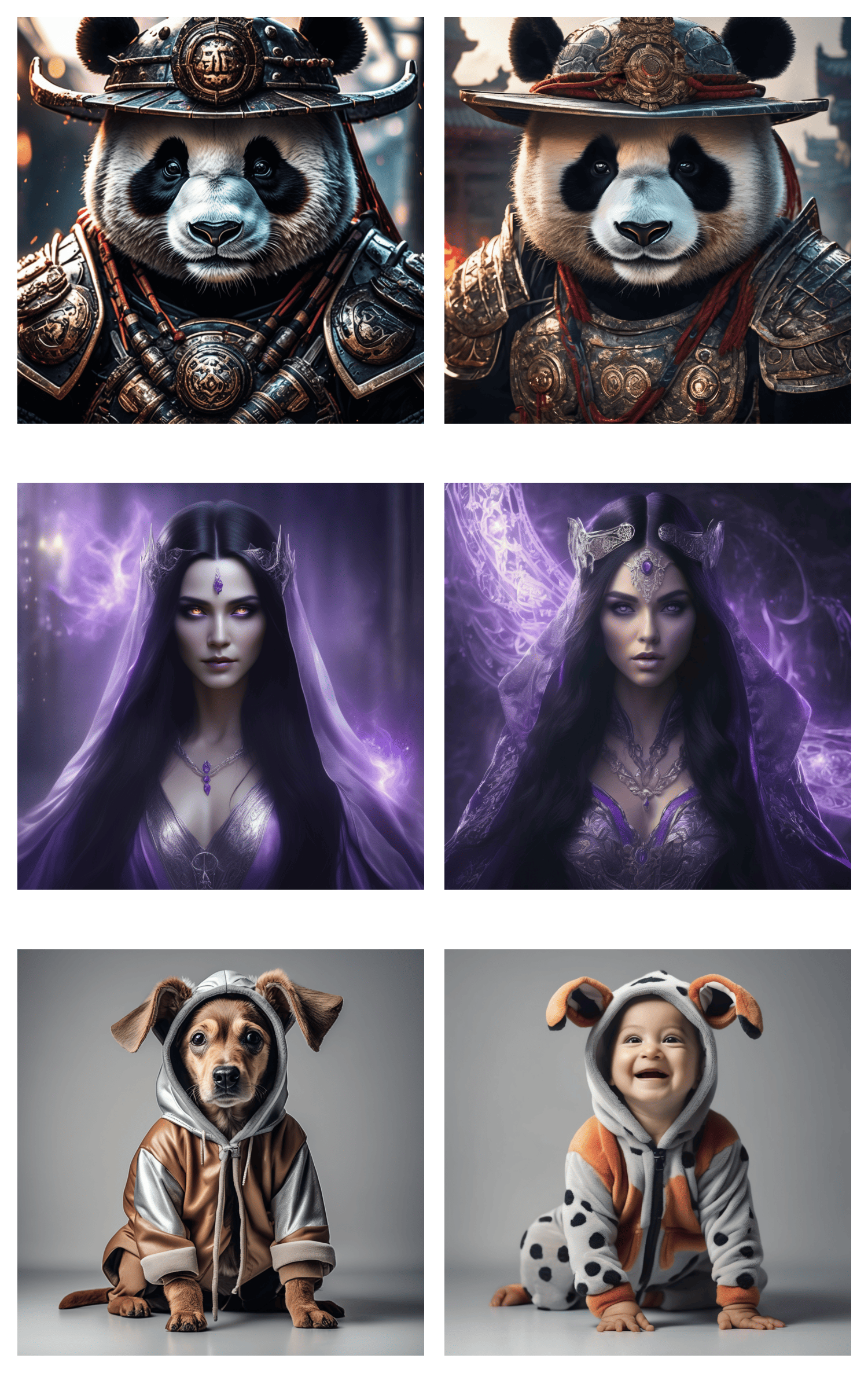}
\caption{\small \textit{SPO} vs \textit{SPO + Ours}}
\end{subfigure}
\hfill
\begin{subfigure}{0.32\textwidth}
\includegraphics[width=\linewidth]{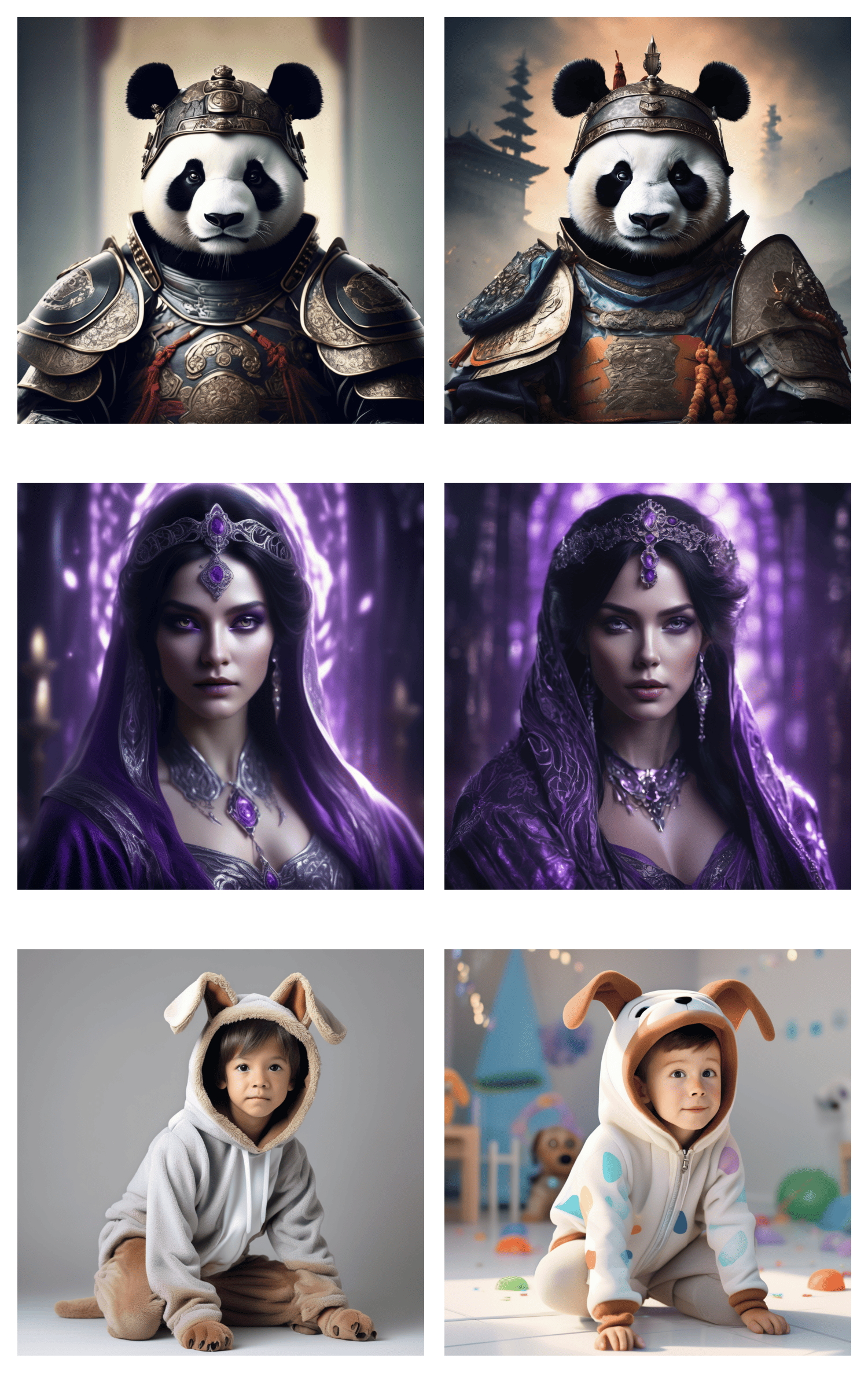}
\caption{\small \textit{Diffusion-DPO} vs \textit{DPO + Ours}}
\end{subfigure}
\caption{\textbf{Plug-in enhancement on SOTA models.} Each pair compares the original outputs (left) and the corresponding results after applying CritiFusion (right). CritiFusion consistently improves semantic alignment and visual fidelity while preserving the original style.}
\label{fig:plugin}
\end{figure*}

\medskip
\noindent\textbf{SDXL based models and plug-in combinations.}
On SDXL (as shown in Table~\ref{tab:sdxl}), CritiFusion again yields substantial improvements.
PickScore increases from 21.97 to 23.13 (+1.16), HPSv2 from 0.266 to 0.291 (+0.025), ImageReward from 0.776 to 1.076 (+0.300), and Aesthetic from 5.94 to 6.22 (+0.28).
Among single model entries, CritiFusion attains the highest HPSv2 (0.291), indicating strong prompt adherence on a high capacity backbone.
Its ImageReward closely matches DyMO (1.076 vs.\ 1.074), while offering comparable Aesthetic.
Although DyMO surpasses CritiFusion in PickScore, CritiFusion remains competitive without relying on reward learning or retraining.

Plug-in variants further highlight complementary behaviors.
Applying CritiFusion to SPO elevates its PickScore from 22.81 to 24.09 (+1.28), HPSv2 from 0.278 to \textbf{0.304}, and ImageReward from 1.082 to \textbf{1.183}, while maintaining Aesthetic in a similar range (6.29, slightly lower than SPO at 6.32).
This combination attains the highest HPSv2 (0.304) and a strong PickScore (24.09), and no other entry simultaneously matches this level of alignment and semantic quality.
Compared to \textit{SPO+DyMO} (23.85 / 0.282 / 1.166), the refinement improves PickScore by +0.24, HPSv2 by +0.022, and ImageReward by +0.017, with no perceptual degradation.

Even high performing baselines such as SD v3.5 and FLUX are outperformed on all four metrics, despite their training level enhancements.
These results collectively validate CritiFusion’s effectiveness and generalizability across diffusion backbones.
Its ability to function as both a standalone sampler and a post hoc refinement stage underscores practical value in scenarios where retraining is costly or infeasible.

\subsection{Qualitative Results}
We present qualitative comparisons (as shown in Figure~\ref{fig:qualitative}) spanning architecture, portraits, animals, indoor still life, birds, and surreal scenes. Across categories, CritiFusion produces images with higher prompt adherence and stronger visual coherence than competing methods. Attribute binding is more reliable for color, material, and accessories, object counts and relations are respected, and boundaries are crisper with fewer halo or ringing artifacts. Global illumination remains consistent with the described scene, avoiding the overly smooth appearance often observed in reward tuned models and the local texture drift sometimes introduced by dynamic guidance. Background structure aligns with the foreground narrative, preserving perspective, occlusion, and depth cues, and low-frequency tone and high-frequency detail are balanced without saturation shifts. These improvements remain consistent across runs and datasets, showing that the gains primarily stem from our critique and spectral modules rather than random variation.

We further evaluate CritiFusion as a plug-in (as shown in Figure~\ref{fig:plugin}). Refinement applied to FLUX, SPO, and Diffusion DPO resolves semantic ambiguities, restores missed attributes, and enhances texture quality while preserving each model’s stylistic signature. Edges are tightened, facial and fur details remain legible at native resolution, metallic and glossy surfaces retain plausible specular highlights, and foliage or fabric displays coherent fine structure. Adjustments are localized and conservative. The original composition is maintained, color casts are corrected rather than replaced, and no spurious content emerges. These observations indicate that CritiFusion functions as a model-agnostic postprocessor that improves alignment and perceptual quality without retraining, complementing both training based and training free baselines.

\subsection{Ablation Study} \label{sec:ablation} 
\begin{figure}[t]
\centering
\includegraphics[width=\linewidth]{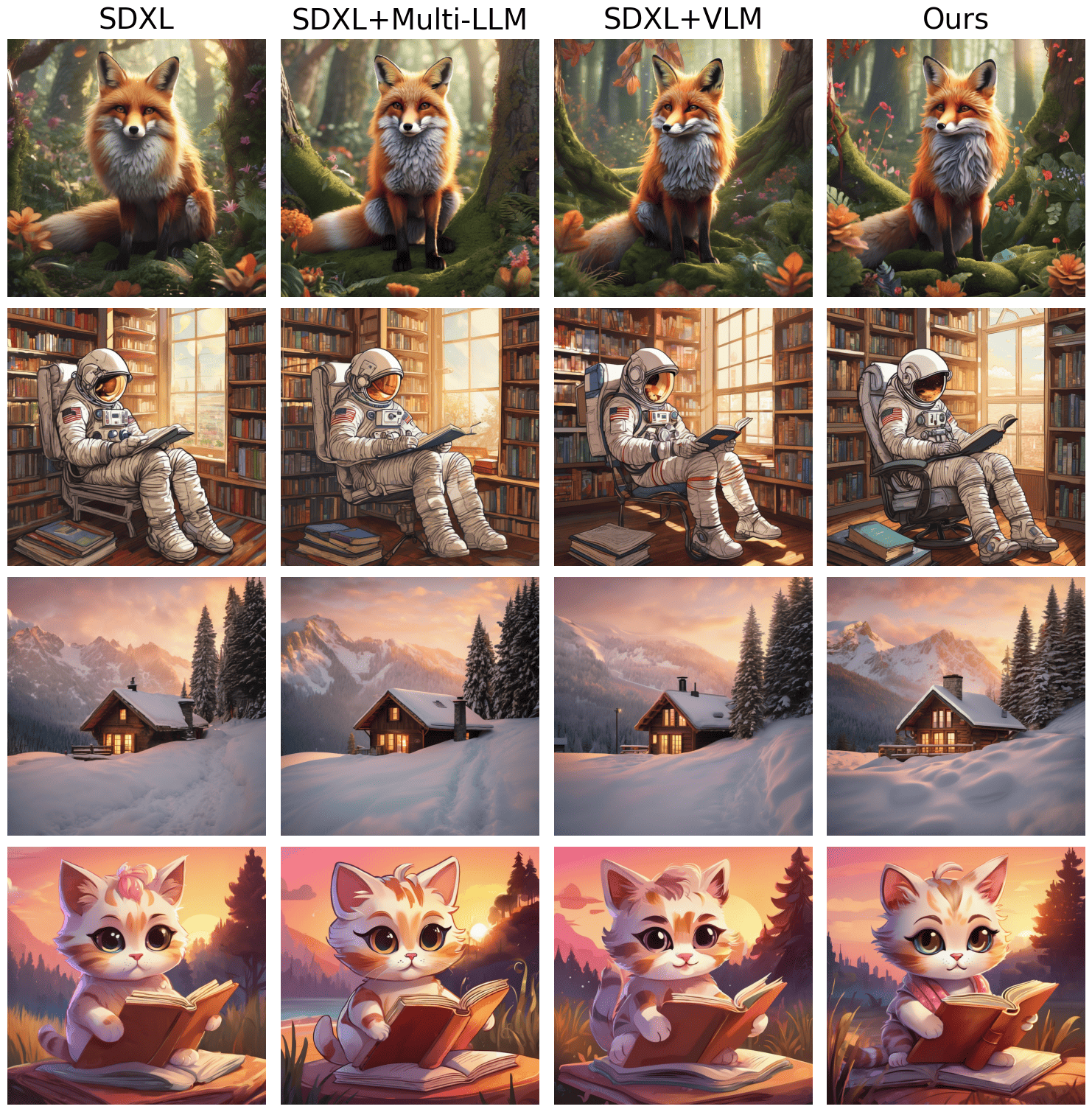}
\caption{\textbf{Qualitative ablations.} Each row shows the outputs from the baseline SDXL, +Multi-LLM, +VLM, and the full CritiFusion model. Adding each component progressively enhances semantic correctness and visual fidelity.}
\label{fig:ablation}
\end{figure}

We conduct two ablations to quantify each module’s contribution in CritiFusion. Table~\ref{tab:ablation} reports a component-wise analysis, Figure~\ref{fig:ablation} provides qualitative evidence, and Table~\ref{tab:llm_count} examines the effect of Multi-LLM ensemble size. All runs use the same SDXL backbone and sampling settings. Only the ablated component or the ensemble size is varied. 

The full system achieves the highest PickScore (23.13) and ImageReward (1.076), with competitive HPSv2 (0.291). Removing image-grounded feedback \textit{(w/o VLM)} yields the largest drops in HPSv2 ($-0.013$) and PickScore ($-0.99$), which highlights the role of visual critique in spatial relations and attribute bindings. Eliminating textual enrichment \textit{(w/o Multi-LLM)} decreases PickScore by $-0.86$, indicating that committee-supplied context recovers fine-grained semantics absent from the raw prompt. Skipping SpecFusion yields a negligible change in alignment (0.293 vs.\ 0.291; $\Delta=0.002$) yet degrades perceptual quality (ImageReward $-0.097$), consistent with SpecFusion’s role in stabilizing global tone and suppressing high-frequency artifacts. Aesthetic varies modestly (6.06--6.22, $\Delta=0.16$), suggesting lower sensitivity than alignment or perceptual quality. 

Varying the committee size (as shown in Table~\ref{tab:llm_count}) produces a monotonic PickScore gain from 22.27 (1~LLM) to 23.13 (5~LLMs, $+0.86$), while HPSv2 fluctuates within 0.282--0.300 and ImageReward trends upward. The largest increase occurs from 1 to 2 ($+0.41$), followed by smaller increments for 3 to 5 ($+0.12$, $+0.15$, $+0.18$). Unless otherwise noted, we report \textit{Ours} with a 5-LLM committee for maximal quality. Qualitatively (as shown in Figure~\ref{fig:ablation}), Multi-LLM improves semantic completeness and attribute fidelity, VLM guidance strengthens spatial consistency and background structure, and the full model with SpecFusion yields sharper textures and more natural illumination.

\begin{table}[tb]
\centering
\caption{Ablation on CritiFusion components.}
\label{tab:ablation}
\begingroup
\footnotesize
\setlength{\tabcolsep}{4pt}
\renewcommand{\arraystretch}{1.05}
\begin{tabular}{@{}lcccc@{}}
\toprule
\textbf{Variant} & \textbf{PickScore↑} & \textbf{HPSv2↑} & \textbf{ImageReward↑} & \textbf{Aesthetic↑} \\
\midrule
w/o VLM         & 22.14 & 0.278 & 0.911 & 6.20 \\
w/o Multi-LLM   & 22.27 & 0.286 & 1.072 & 6.14 \\
w/o SpecFusion  & 22.31 & 0.293 & 0.979 & 6.06 \\

\textbf{Ours}   & \textbf{23.13} & \textbf{0.291} & \textbf{1.076} & \textbf{6.22} \\
\bottomrule
\end{tabular}
\endgroup
\end{table}

\begin{table}[tb]
\centering
\caption{Effect of LLM ensemble size on CritiFusion performance.}
\label{tab:llm_count}
\begingroup
\footnotesize
\setlength{\tabcolsep}{6pt}
\renewcommand{\arraystretch}{1.05}
\begin{tabular}{@{}ccccc@{}}
\toprule
\textbf{\# LLMs} & \textbf{PickScore↑} & \textbf{HPSv2↑} & \textbf{ImageReward↑} & \textbf{Aesthetic↑} \\
\midrule
1 & 22.27 & 0.288 & 0.972 & 6.16 \\
2 & 22.68 & 0.282 & 1.005 & 6.19 \\
3 & 22.80 & 0.297 & 1.010 & 6.13 \\ 
4 & 22.95 & 0.300 & 1.051 & 6.17 \\ 
\textbf{5} & \textbf{23.13} & \textbf{0.291} & \textbf{1.076} & \textbf{6.22} \\
\bottomrule
\end{tabular}
\endgroup
\end{table}

\section{Conclusion}
\noindent
We presented CritiFusion, an inference-time framework that enhances text-to-image diffusion generation through semantic critique and spectral alignment. By integrating external knowledge from vision-language and large language models, our method injects high-level semantic guidance into existing diffusion pipelines without additional training or architecture modification. CritiFusion is model-agnostic and can serve as a plug-in module for a wide range of generators.

Empirical evaluations indicate that CritiFusion strikes a strong balance between prompt faithfulness and visual appeal, remaining competitive with leading fine-tuned systems while requiring no retraining. When combined with other models, CritiFusion further amplifies alignment and perceptual quality, highlighting its complementary nature and scalability.

The effectiveness of CritiFusion opens several avenues for future exploration. Advancing Multi-LLM strategies or iterative feedback loops could refine the critique process, while extending spectral alignment to multi-scale or cross-domain settings may further improve structural and detail preservation. Beyond image synthesis, applying the same inference-time refinement principles to text-to-video or 3D generation presents an exciting opportunity for broader multimodal adaptation.

\appendix
\twocolumn[{
\begin{center}
    {\Huge \textbf{Appendix}} \\[12pt]
    
    {\Large CritiFusion: Semantic and Spectral Refinement for Text-to-Image Diffusion}
\end{center}
\vspace{15mm} 
}]

\section{Reproducibility and Code Release}
\label{app:c-reproducibility}

To promote transparent and verifiable research, we provide an anonymized repository containing the complete implementation of \textbf{CritiFusion}. As CritiFusion operates as a training-free inference-time refinement framework, the provided code focuses on the inference pipeline rather than training scripts. The repository enables the reproduction of our semantic alignment and spectral fusion results using the exact configurations reported in the paper.

The released materials include:
\begin{itemize}
    \item \textbf{Inference Workflow:} The core pipeline is encapsulated in \texttt{critifusion.ipynb}, which guides users through the complete process: base image generation, multi-LLM concept decomposition, VLM critique, and frequency-domain fusion (SpecFusion).
    \item \textbf{Environment \& Dependencies:} We provide \texttt{requirements.txt} for easy environment setup. The code is configured to interface with the Together AI API for accessing the necessary Large Language Models (e.g., Llama-3.3, Qwen-2.5) and Vision Language Models required by the CritiCore module.

\end{itemize}

The code is modular, allowing for independent verification of the semantic and spectral components. The anonymized repository is publicly available at the following link:

\noindent
\textbf{\url{https://github.com/Rossi-Laboratory/CritiFusion}}

This link will remain active to ensure consistent access for verification and reproducibility assessment.

\section{Discussion}
\label{app:Discussion}

\subsection{Limitations}
\label{app:Limitations}
CritiFusion improves semantic alignment and perceptual quality across multiple diffusion backbones, yet the method presents several limitations that follow from its design. The framework depends on external vision-language and large language models for critique and semantic enrichment, and the effectiveness of these signals varies with prompt clarity and the reliability of the upstream models. When prompts are ambiguous or overly broad, the critique may introduce extraneous edits and the aggregator must aggressively prune low-salience phrases to preserve a compact conditioning. The method also introduces additional computation because each generation requires a corrective img2img pass followed by spectral fusion. Although the overhead is moderate, the overall cost is higher than single-pass sampling and becomes more noticeable when operating large backbones or evaluating many prompts. The approach assumes that the vision-language model provides stable clause-level scores, and misidentification of attributes or relations can propagate into the refinement and lead to unnecessary adjustments. SpecFusion is designed to preserve global structure through low-frequency retention, which limits the degree of structural change that can be introduced during refinement. Base generations with severe layout or geometry errors may therefore remain only partially corrected. The method is further designed for single-image synthesis and does not provide explicit mechanisms for temporal consistency or multi-view coherence. Extending the approach to video or 3D generation will require additional modeling to maintain stability across frames or viewpoints.

\subsection{Societal Impact}
\label{app:Societal Impact}
CritiFusion enhances the semantic reliability and visual quality of text-to-image diffusion models, which can broaden the accessibility of high-fidelity content creation for artists, educators, designers, and general users. At the same time, improved alignment and realism may amplify existing concerns in generative media. The system can produce highly convincing outputs that resemble authentic scenes, which raises the possibility of misuse in misinformation, impersonation, or deceptive visual communication. Because CritiFusion operates as a training-free refinement stage and can be attached to a wide range of diffusion backbones, the method may lower the barrier for generating persuasive synthetic imagery. Careful deployment is therefore necessary, including transparency about synthetic content and responsible use guidelines.

The framework depends on external vision-language and language models whose responses may reflect demographic, cultural, or data-driven biases. These signals influence the refinement process and can propagate implicit biases into the final images. Future work may benefit from bias analysis tools, calibrated critique mechanisms, and broader evaluations on diverse prompts to mitigate unintended cultural or representational skew. Although CritiFusion improves prompt adherence, the method does not include explicit constraints for safety filtering or content moderation. Applications in sensitive domains should integrate appropriate safeguards to ensure that refinement does not unintentionally amplify harmful or inappropriate content. Overall, CritiFusion provides practical benefits for creative workflows, but responsible usage and awareness of ethical considerations remain essential for minimizing potential negative societal impact.

\section{Preliminaries}
\label{app:preliminaries}

\subsection{Diffusion and SDXL}
\label{app:a-ldm-sdxl}

\subsubsection{LDM Preliminaries} 
\label{app:a-ldm}
\paragraph{Forward noising.}
Let $\{\alpha_t\}_{t=1}^{T}$ be a variance schedule and $\bar{\alpha}_t=\prod_{s=1}^{t}\alpha_s$.
A clean latent $\mathbf{x}_0$ is corrupted via
\begin{equation}
q(\mathbf{x}_t \mid \mathbf{x}_0)
=\mathcal{N}\!\bigl(\mathbf{x}_t;\, \sqrt{\bar{\alpha}_t}\,\mathbf{x}_0,\; (1-\bar{\alpha}_t)\mathbf{I}\bigr).
\label{eq:forward}
\end{equation}

\paragraph{Classifier-free guidance (CFG).}
Given prompt $c$ and null condition $\emptyset$, the guided noise readout is
\begin{equation}
\widehat{\epsilon}_\theta(\mathbf{x}_t,c;w)
=(1{+}w)\,\epsilon_\theta(\mathbf{x}_t,c)-w\,\epsilon_\theta(\mathbf{x}_t,\emptyset),
\label{eq:cfg}
\end{equation}
with guidance scale $w\!\ge\!0$.
An $\mathbf{x}_0$ estimate follows
\begin{equation}
\widehat{\mathbf{x}}_0
=\frac{\mathbf{x}_t-\sqrt{1-\bar{\alpha}_t}\,\widehat{\epsilon}_\theta(\mathbf{x}_t,c;w)}{\sqrt{\bar{\alpha}_t}}.
\label{eq:x0-pred}
\end{equation}

\paragraph{DDIM update.}
For $\eta{=}0$,
\begin{equation}
\mathbf{x}_{t-1}
=\sqrt{\bar{\alpha}_{t-1}}\,\widehat{\mathbf{x}}_0
+ \sqrt{1-\bar{\alpha}_{t-1}}\,\widehat{\epsilon}_\theta(\mathbf{x}_t,c;w).
\label{eq:ddim}
\end{equation}

\paragraph{High-order solvers.}
We use DPM-Solver++ (order 2/3) with a Karras $\sigma$ discretization for better quality–step tradeoffs; this choice is orthogonal to $w$ and to our two-pass usage.

\paragraph{Latent-space autoencoding.}
All denoising runs in a VAE latent space with scale $\gamma>0$.
\begin{equation}
\mathbf{z}=\gamma\,\mathrm{VAE}^{\text{enc}}(\mathbf{x}),\;
\widehat{\mathbf{x}}=\mathrm{VAE}^{\text{dec}}\!\Big(\tfrac{\mathbf{z}}{\gamma}\Big).
\label{eq:vae-scale}
\end{equation}
We adopt $\gamma\!=\!0.18215$ for SD\,1.5 and $\gamma\!=\!0.13025$ for SDXL.

\subsubsection{SDXL Two-Pass Generation} 
\label{app:a-sdxl}
\paragraph{Backbone and text conditioning.}
SDXL scales the UNet and uses two text encoders (e.g., CLIP ViT-L/14 and OpenCLIP ViT-bigG/14) with wider cross-attention, targeting native $1024{\times}1024$ latents. A dedicated refiner is commonly used for late steps.

\paragraph{Base vs.\ corrective passes.}
We separate (i) a \emph{base} pass that samples from noise using the prompt, and (ii) a short \emph{corrective} pass that performs latent img2img (SDEdit-style) on the base latents to sharpen details while keeping layout. We \emph{fix the random seed} across both passes to isolate sampler/text effects from randomness.

\paragraph{Strength--step mapping (img2img).}
Let $T'$ be corrective steps and ``last-$k$'' denote replacing the first $T'{-}k$ steps by re-noising the input:
\begin{align}
s_{\text{img2img}} &= \mathrm{clip}\!\Big(\tfrac{k}{T'},\,0.01,\,0.95\Big), \nonumber \\
t_0 &= \big\lfloor (1-s_{\text{img2img}})\,T' \big\rfloor,
\label{eq:strength}
\end{align}
where $s_{\text{img2img}}$ is \texttt{strength} and $t_0$ is the denoising start. Smaller $s_{\text{img2img}}$ preserves more low frequencies from the base pass.

\paragraph{Variance matching.}
Decoding with the model-correct $\gamma$ in~\eqref{eq:vae-scale} avoids washed-out or over-saturated outputs when mixing SD\,1.5 and SDXL components.

\subsection{Multi-Agent Framework}
\label{app:b-mad-moa}

\subsubsection{Multi-Agent Debate (MAD)} 
\label{app:b-mad}
\paragraph{Setup.}
Given instruction $x$, we instantiate $M$ agents $\mathcal{A}=\{f_i\}_{i=1}^{M}$ and run $T$ rounds. At round $t$,
\begin{align}
y_i^{(t)} &= f_i\!\big(x,\; \mathrm{ctx}^{(t-1)}_{\neg i}\big), \nonumber \\
\mathrm{ctx}^{(t-1)}_{\neg i} &= \{(j,\,y_j^{(t-1)}): j\neq i\}.
\label{eq:mad-update}
\end{align}
A judge $\mathcal{J}$ selects or synthesizes the final answer
\begin{equation}
\hat{y}\;=\;\mathcal{J}\!\left(x,\{y_i^{(T)}\}_{i=1}^{M}\right).
\label{eq:mad-agg}
\end{equation}
We enforce token budgets and semantic de-duplication to keep outputs concise and checkable.

\paragraph{Design choices.}
We favor heterogeneous agents (diversity), short debates ($T\in\{1,2\}$) for latency, and a strong judge. Unsafe or image-inconsistent edits flagged by VLM critics are discarded.

\subsubsection{Mixture-of-Agents (MoA)} 
\label{app:b-moa}
\paragraph{Layered committee.}
MoA organizes collaboration into $L$ layers. Layer $l$ has $n_l$ proposers $\{g_{l,j}\}_{j=1}^{n_l}$ and an aggregator $A_l$:
\begin{equation}
r_{l,j} = g_{l,j}\!\big([x;\, s_{l-1}] \big), \quad
s_l = A_l\!\big(x,\{r_{l,j}\}_{i=1}^{n_l}\big),
\label{eq:moa}
\end{equation}
with $s_0{=}x$ and final output $\hat{y}=s_L$. Proposers contribute diverse candidates; the aggregator \emph{aggregates-and-synthesizes} rather than merely ranking.

\paragraph{Complexity and budget.}
A $T$-round MAD uses $\mathcal{O}(M\,T)$ calls plus judging, whereas an $L$-layer MoA uses $\sum_{l=1}^{L}(n_l{+}1)$ calls, often parallelizable across proposers. Caching intermediate candidates reduces end-to-end latency.

\paragraph{Pipeline integration.}
We use the consolidated text to drive a short img2img corrective pass under fixed seeds, then apply our frequency-domain fusion; this yields stable alignment gains without re-sampling from pure noise.

\section{More Details of the Method}
\label{app:method-details}

\subsection{Controlled Adaptive Denoising Rate}
\label{app:d-cadr}

The Controlled Adaptive Denoising Rate (CADR) maps an alignment confidence score
$s\!\in[0,1]$ from the VLM critique or a preference predictor to the
hyperparameters of the corrective img2img pass.
Following Eq.~(2) of the main paper,
\[
(\lambda, g, T', \rho) = \Phi_{\mathrm{CADR}}(s) = A + (1-s)B ,
\]
where $\lambda$ denotes the img2img strength, $g$ the guidance scale,
$T'$ the number of refinement steps, and $\rho$ the SpecFusion frequency-mask ratio.
Lower confidence triggers stronger correction while preserving low-frequency structure.

In practice CADR is a lightweight affine interpolation:
\begin{align}
\lambda &= \mathrm{lerp}(0.12,\,0.30,\,1-s), \nonumber \\
g &= \mathrm{lerp}(3.6,\,5.0,\,1-s), \nonumber \\
T' &= \mathrm{round}\!\big[\mathrm{lerp}(16,\,30,\,1-s)\big], \nonumber \\
\rho &= \mathrm{lerp}(0.60,\,0.85,\,1-s).
\label{eq:cadr-mapping}
\end{align}
This is implemented as \texttt{cadr\_from\_alignment()} in our code.
Refinement is skipped when $s\!>\!0.9$.
CADR adds negligible overhead since the corrective pass uses at most $30$ steps.

\subsection{Img2img step sweep} \label{app:k-sweep} 

\begin{figure}[t]
\centering
\includegraphics[width=\linewidth]{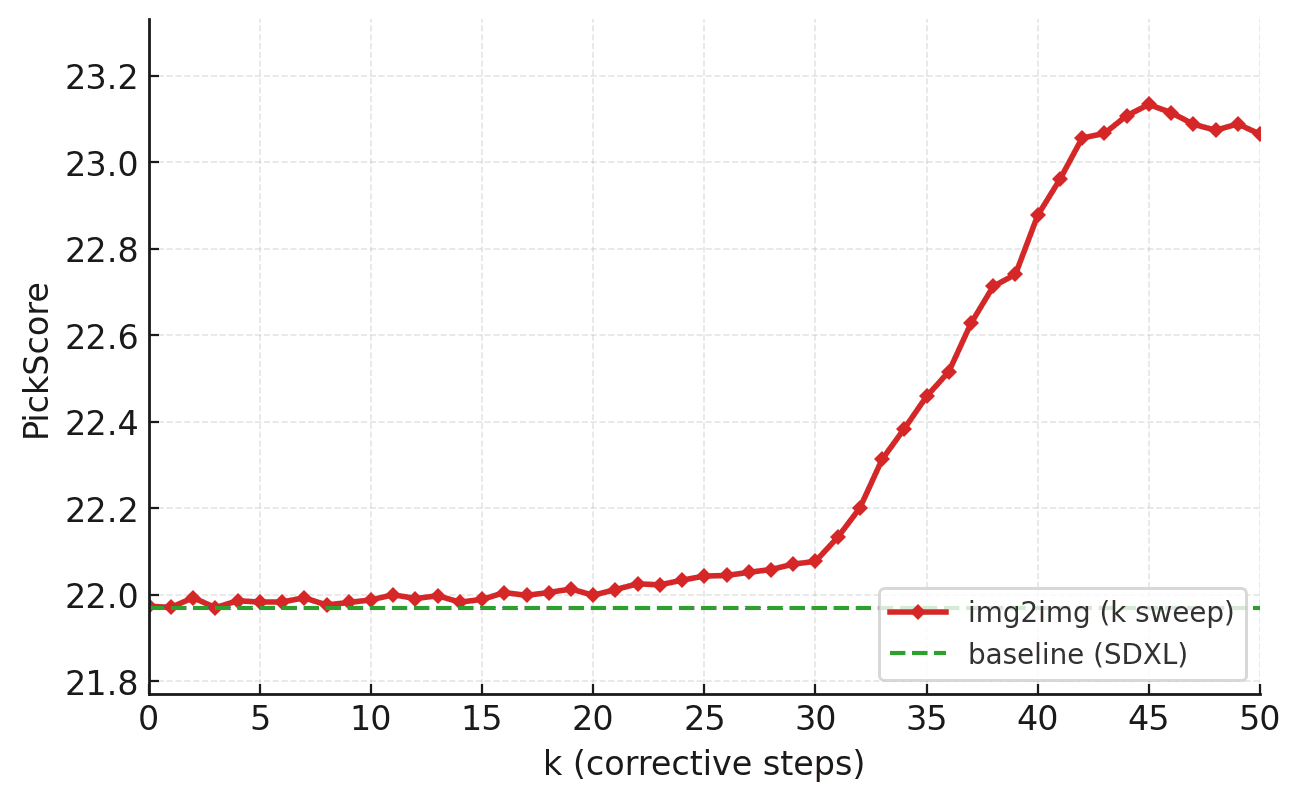}
\vspace{-6pt}
\caption{\textbf{PickScore vs.\ corrective steps $k$.}}
\label{fig:app_k_pick}
\vspace{-6pt}
\end{figure}

\begin{figure}[t]
\centering
\includegraphics[width=\linewidth]{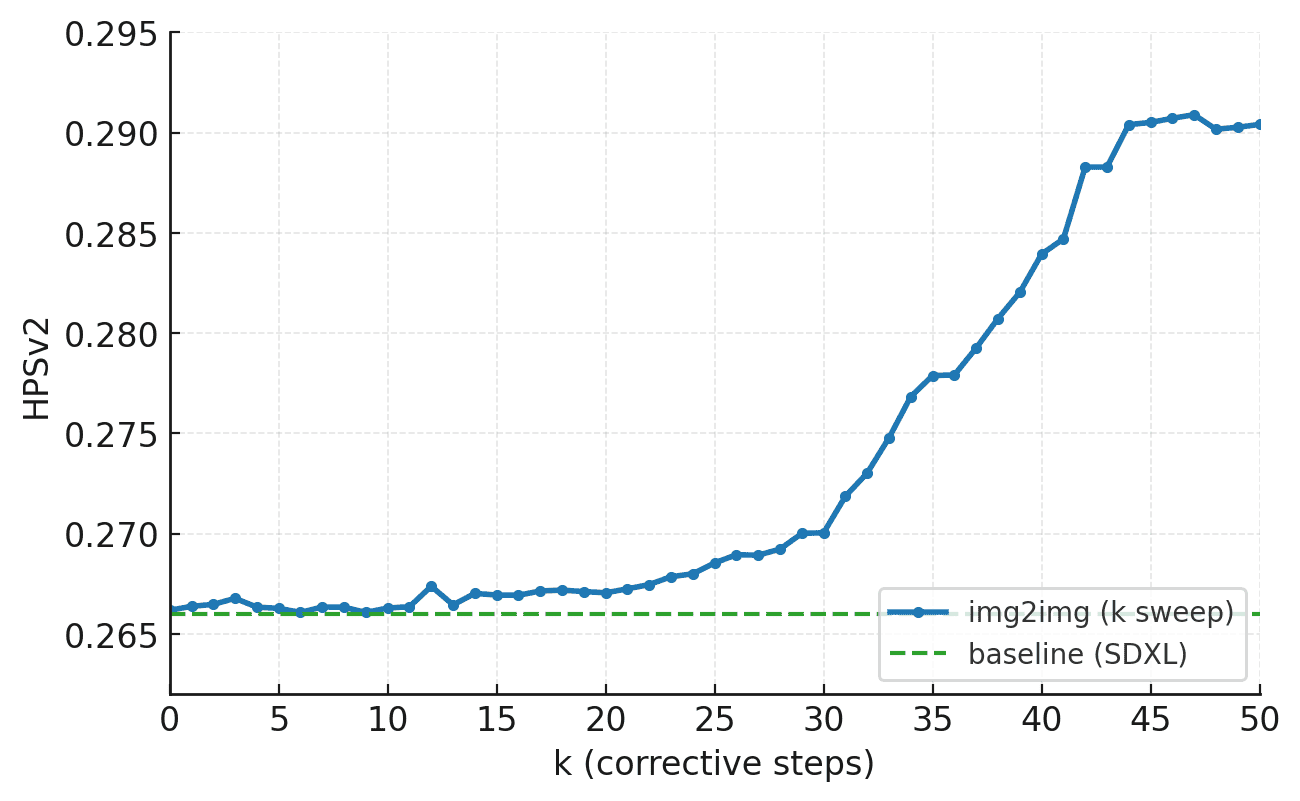}
\vspace{-6pt}
\caption{\textbf{HPSv2 vs.\ corrective steps $k$.}}
\label{fig:app_k_hps}
\vspace{-6pt}
\end{figure}

\begin{figure*}[t]
  \centering
  \includegraphics[width=\textwidth,height=18cm]{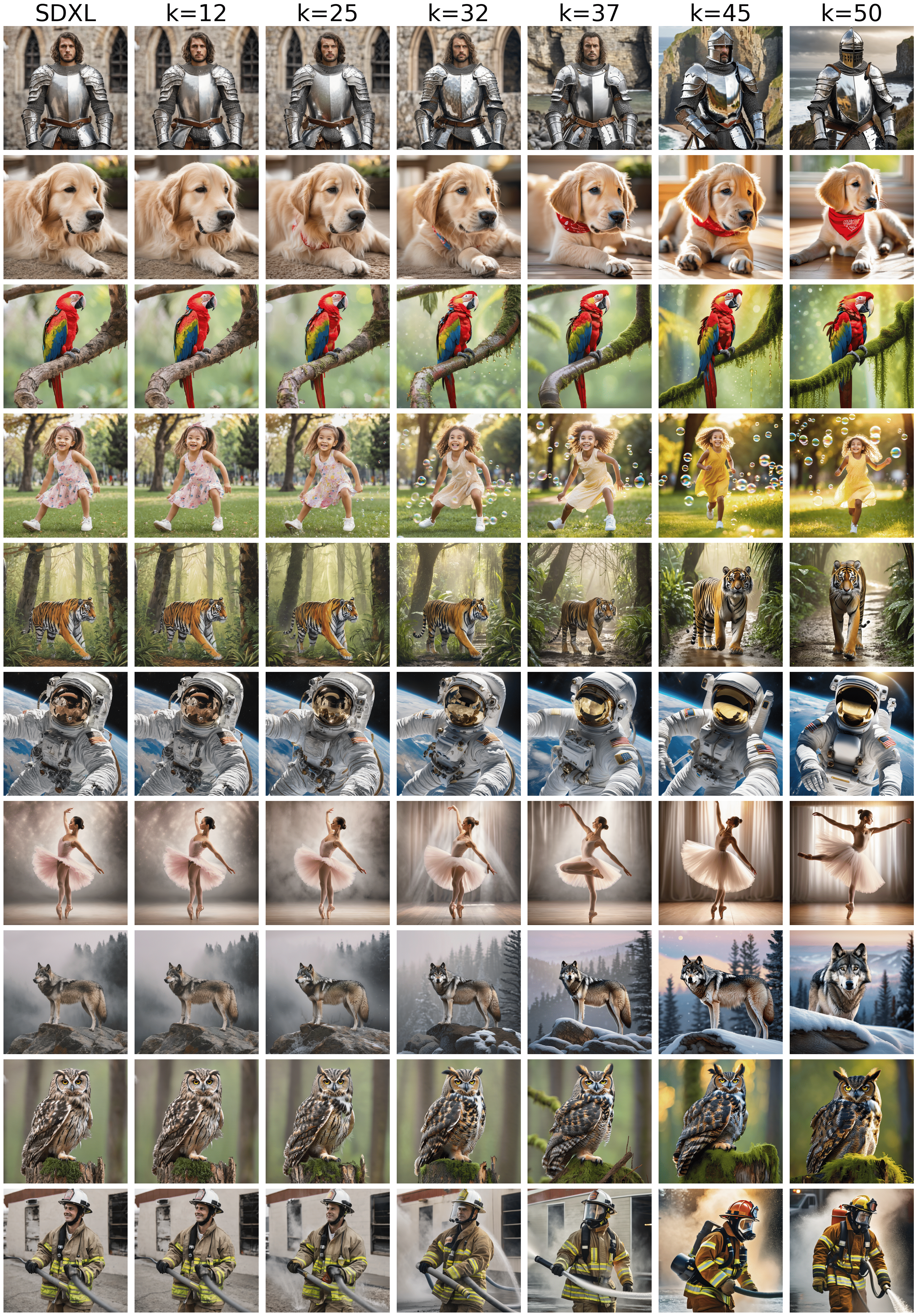}
  \vspace{-8pt}
  \caption{\textbf{Qualitative progression across $k$.}
  From left to right: SDXL base and img2img with $k \in \{12, 25, 32, 37, 45, 50\}$. The corrective pass stabilizes global layout while gradually improving local details and attribute faithfulness.}
  \label{fig3}
  \vspace{-8pt}
\end{figure*}

\noindent\textbf{Rationale.} We investigate how the number of corrective diffusion steps $k$ in the img2img refinement pass influences alignment accuracy, semantic preference, and overall perceived quality. The corrective pass can be interpreted as a controlled noise-injection mechanism that selectively adjusts high-frequency structure while preserving the low-frequency scene layout. As visualized in Figure~\ref{fig:app_k_pick} and Figure~\ref{fig:app_k_hps}, both PickScore and HPSv2 remain close to the SDXL baseline when $k$ is small ($0 \le k \le 30$), indicating that early iterations provide only marginal deviation from the initial sample. Once $k$ enters the intermediate regime ($31 \le k \le 40$), both metrics exhibit a clear upward trend, suggesting that moderate corrective strength unlocks meaningful refinements without destabilizing scene composition. The performance peak consistently emerges around $k\!\approx\!45$ , forming a narrow but stable plateau from $k=44$ to $k=46$. Beyond this window, the curves remain comparatively high yet display small oscillations, and the setting around $k=50$ is usually slightly below the optimum. Figure~\ref{fig3} qualitatively illustrates how increasing k enhances object-level fidelity, fine-grained attributes, and local geometric consistency while maintaining overall scene layout.

\noindent\textbf{Protocol.} To ensure a controlled and reproducible evaluation, all curves are computed over a fixed and diverse prompt set that includes portraits, object-centric prompts, and full-scene descriptions. The seeds, scheduler, guidance-scale floor, and negative prompts are held constant so that the only experimental variable is the corrective-step parameter $k$. We use PickScore and HPSv2 because they capture complementary notions of user preference, semantic alignment, and instruction fidelity. Each prompt is truncated to a 77-token limit to minimize variation caused by prompt-length differences. All hyperparameters and sampling conditions are synchronized across the entire experiment. The dashed reference line in the plots corresponds to the SDXL baseline without any corrective refinement, which helps isolate the effect of varying k and clarifies how corrective strength influences structural preservation and detail recovery.

\noindent\textbf{Observations.} Both PickScore and HPSv2 display a clear upward trend once $k$ is moderately large, which begins around a value of thirty. Low values of k preserve the global layout but provide limited high-frequency refinement. As $k$ increases, the refinement path receives enough controlled noise to regenerate sharp textures, resolve attribute mismatches, and improve semantic coherence, while the underlying composition remains intact. Very large values of $k$ introduce a small amount of overcorrection. This manifests as excessively sharpened details, local inconsistencies, or texture drift, and accounts for the soft decrease after the maximum near $k\!\approx\!45$. The qualitative progression presented in Figure~\ref{fig3} follows the same pattern. Increasing k consistently strengthens local detail while the large-scale structure remains stable, which confirms that corrective refinement has a predictable influence on the visual outcome.

\noindent\textbf{Practical choice.} Experimental evidence indicates that the range from $k = 44$ to $k = 46$ provides the strongest overall performance. This region balances fidelity improvements, computational efficiency, and scene consistency. It introduces sufficient refinement to recover crisp edge structure and accurate attributes while avoiding the noise amplification that accompanies extremely high settings. In typical workflows, $k$ is implemented as the number of final steps in an img2img schedule. Larger values correspond to stronger noise injection during the refinement pass. Settings around $k = 45$ supply enough stochasticity to enhance high-frequency detail, yet they maintain the low-frequency geometry inherited from the base sample. For practitioners seeking reliable and high-quality results, this region offers an effective and robust default.

\section{Additional Render Results}
\label{app:render_results}
\begin{figure*}[t]
  \centering
  \includegraphics[width=\textwidth, height=18cm]{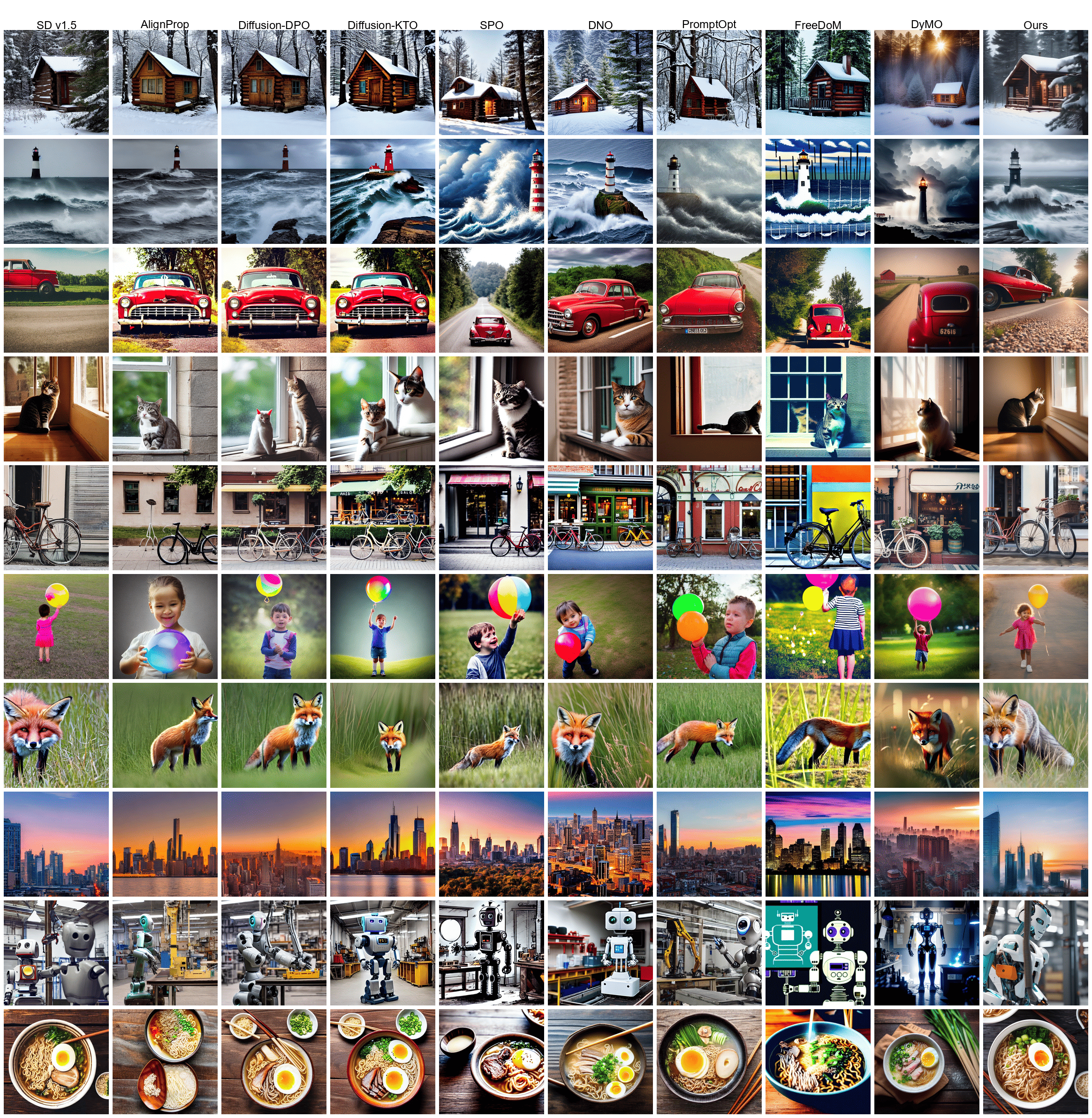} 
  \vspace{-8pt}
  \caption{\textbf{Qualitative comparison on SDv1.5-based methods} }

  \label{fig4} 
  \vspace{-6pt}
\end{figure*}

\begin{figure*}[t]
  \centering
  \includegraphics[width=\textwidth, height=18cm]{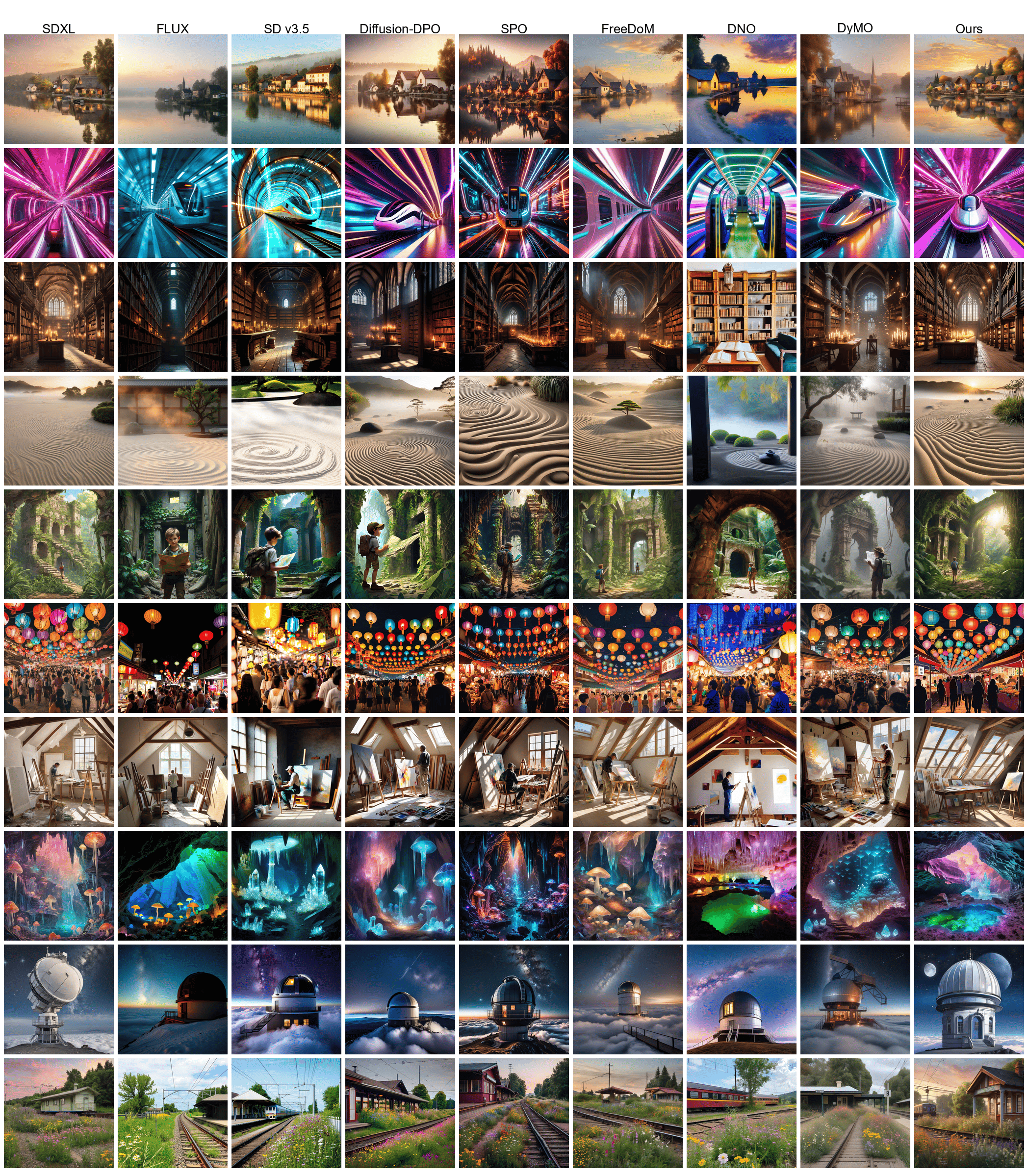} 
  \vspace{-8pt}
  \caption{\textbf{Qualitative comparison on SDXL-based methods} }

  \label{fig5} 
  \vspace{-6pt}
\end{figure*}

\begin{figure*}[t]
  \centering
  \includegraphics[width=\textwidth, height=18cm]{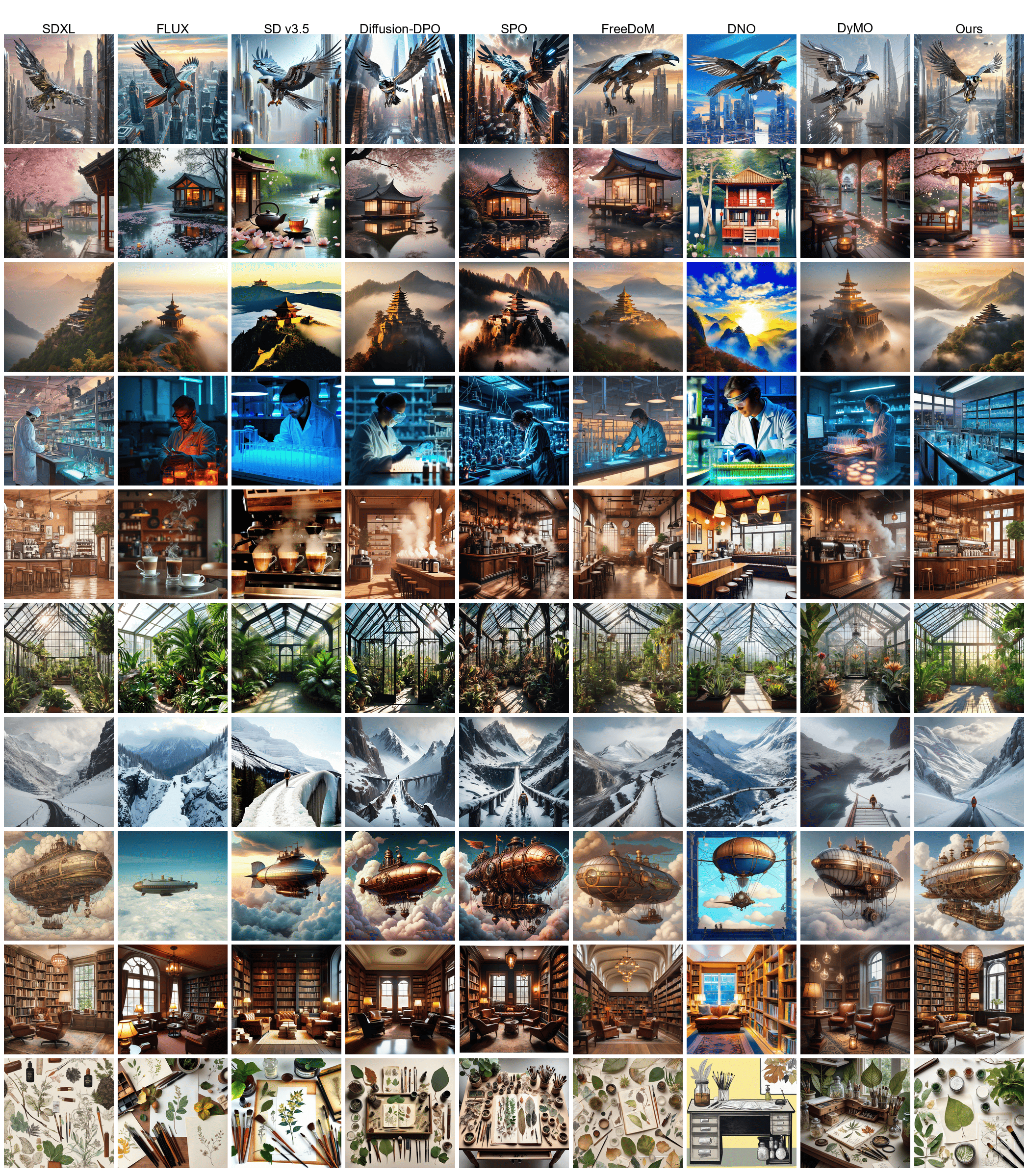} 
  \vspace{-8pt}
  \caption{\textbf{Qualitative comparison on SDXL-based methods} }
  \label{fig6} 
  \vspace{-6pt}
\end{figure*}

A broader examination of visual outputs across a wide range of prompts shows clear differences in how existing SD v1.5 based methods handle detail, structure, and prompt fidelity. As illustrated in Figures ~\ref{fig4}, ~\ref{fig5}, and~\ref{fig6}, baseline approaches frequently lose high frequency information, exhibit attribute inconsistencies in color or material, or generate unstable illumination and spatial configuration. These issues often emerge when prompts involve multiple entities, complex compositions, fine textures, or subtle relationships that require stable cross attention behavior. In scenes with reflective materials, curved surfaces, layered backgrounds, or delicate structural patterns, baseline outputs often appear washed out or internally inconsistent. Our refinement method effectively addresses these weaknesses by maintaining stable global composition while strengthening local detail and semantic correctness. Textures remain crisp even in small regions, object boundaries stay well defined under heavy occlusion, and material or color attributes follow the intended description with improved reliability. The improvements appear consistently across vehicles, animals, portraits, indoor environments, and stylized scenes, where the refined results display more coherent lighting, cleaner depth transitions, and more faithful interpretation of prompt specified attributes. Additional examination also shows that the refined images exhibit fewer corner artifacts, more stable shading gradients, and reduced incidence of unwanted hallucinated elements. These properties contribute to an overall improvement in realism and readability, especially in challenging scenes that combine both natural and human made structures.

\begin{figure*}[t]
  \centering
  \includegraphics[width=\textwidth, height=18cm]{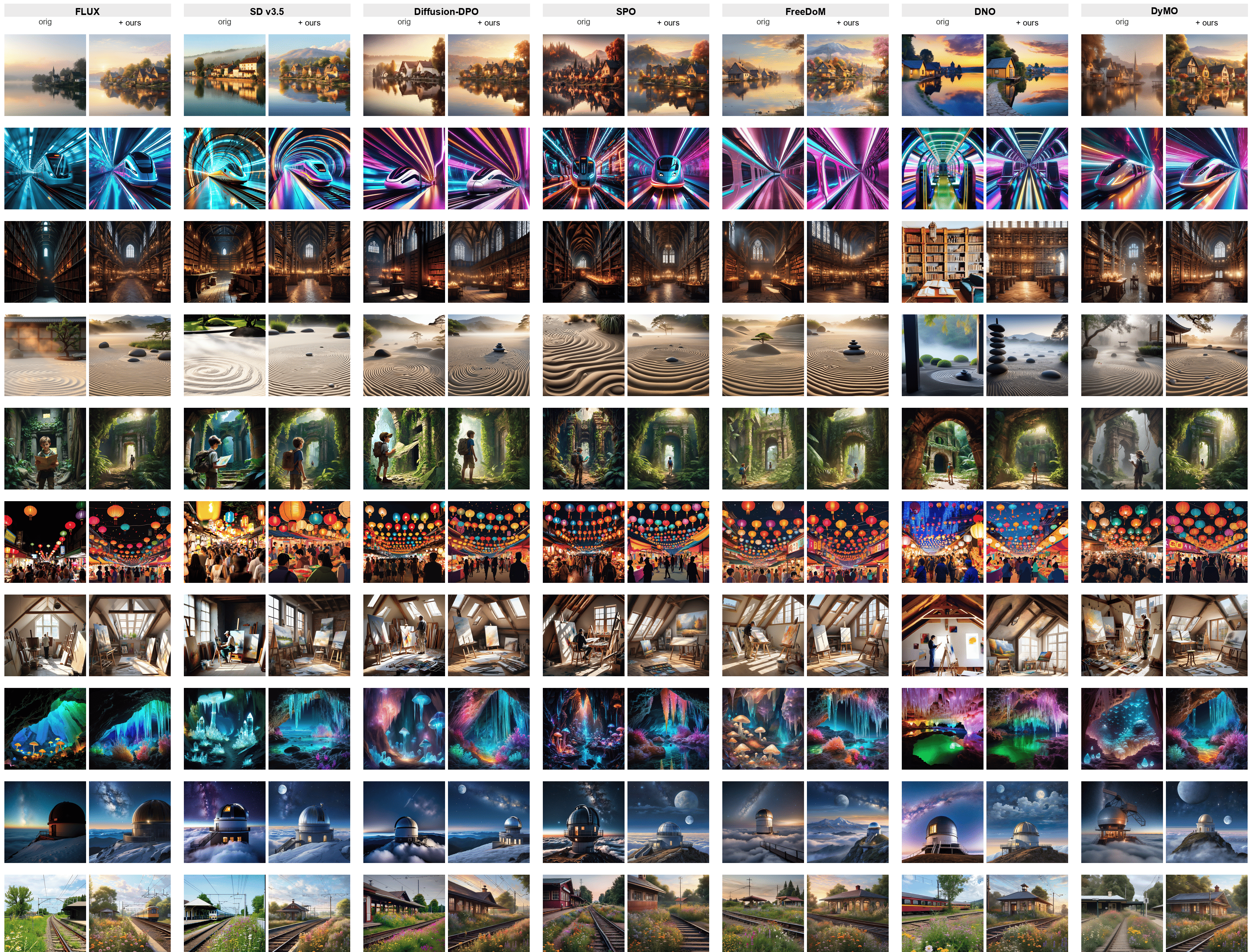} 
  \vspace{-8pt}
  \caption{\textbf{Cross-model refinement consistency across diverse generative backbones.} }
  \label{fig7} 
  \vspace{-6pt}
\end{figure*}

\begin{figure*}[t]
  \centering
  \includegraphics[width=\linewidth, height=18cm]{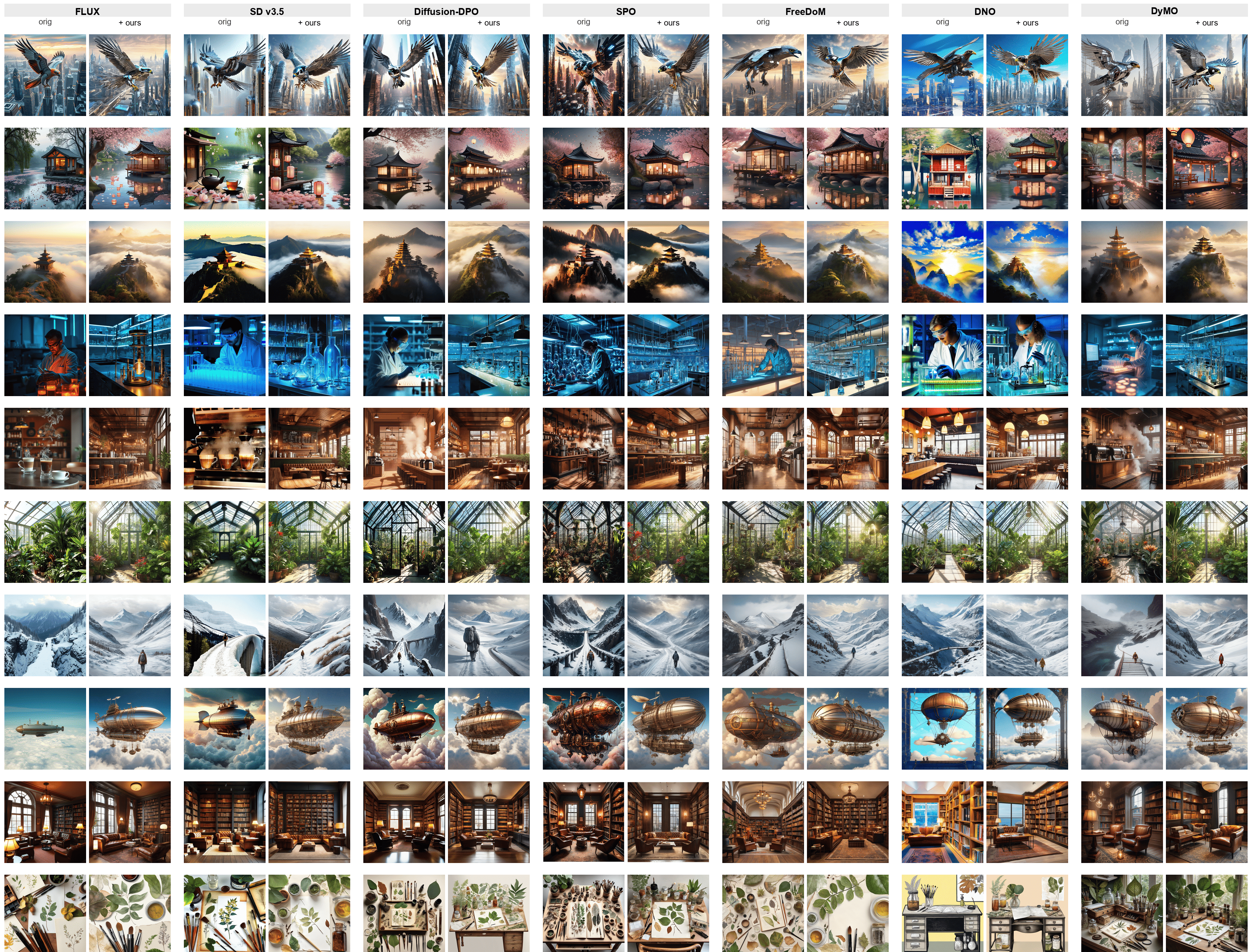}
  \vspace{-6pt}
  \caption{\textbf{Cross-model refinement consistency across diverse generative backbones}}
  \label{fig8}
  \vspace{-4pt}

\end{figure*}

Further evidence of generalization is visible in Figures ~\ref{fig7} and ~\ref{fig8}, where the refinement stage is applied to a diverse set of generative backbones that include FLUX, SD v3.5, Diffusion DPO, SPO, FreeDoM, DNO, and DyMO. In each model, the original generations often contain softened surfaces, irregular reflections, incomplete structural elements, or localized distortions when handling difficult compositions. After refinement, the outputs recover sharper high frequency content, more coherent illumination, and more stable geometry while preserving the unique visual style associated with each backbone. The improvements extend across all examined categories, including natural landscapes, high speed transportation scenes, large indoor architectures, stylized zen environments, and dense natural settings. These observations indicate that the proposed refinement approach enhances clarity, consistency, and scene fidelity in a model agnostic manner and provides reliable benefits across both compositional and stylistically diverse prompts.

{\small
\begin{thebibliography}{30}

\bibitem{dymo}
X.~Xie and D.~Gong.
DyMO: training-free diffusion model alignment with dynamic multi-objective scheduling.
In \textit{Proceedings of the IEEE/CVF Conference on Computer Vision and Pattern Recognition}, 2025.

\bibitem{laion}
C.~Schuhmann.
LAION-Aesthetics: aesthetic score predictor v2.
LAION Blog, 2022. [Accessed: 2025-11-11].

\bibitem{clip}
A.~Radford, J.~W. Kim, C.~Hallacy, A.~Ramesh, G.~Goh, S.~Agarwal, G.~Sastry, A.~Askell, P.~Mishkin, J.~Clark, G.~Krueger, and I.~Sutskever.
Learning transferable visual models from natural language supervision.
In \textit{Proceedings of the International Conference on Machine Learning}, 2021.

\bibitem{gpt4}
OpenAI.
GPT-4 technical report.
\textit{arXiv preprint arXiv:2303.08774}, 2023.

\bibitem{parti}
J.~Yu, Y.~Xu, J.~Y. Koh, T.~Luong, G.~Baid, Z.~Wang, V.~Vasudevan, A.~Ku, Y.~Yang, B.~K. Ayan, et al.
Scaling autoregressive models for content-rich text-to-image generation.
\textit{Transactions on Machine Learning Research}, 2022.

\bibitem{hugginggpt}
Y.~Shen, K.~Liu, Z.~Zhang, J.~Cheng, K.~Zhao, D.~Zhao, and W.~Chen.
HuggingGPT: solving AI tasks with ChatGPT and its friends in Hugging Face.
\textit{arXiv preprint arXiv:2303.17580}, 2023.

\bibitem{instructgpt}
L.~Ouyang, J.~Wu, X.~Jiang, D.~Almeida, C.~Wainwright, P.~Mishkin, C.~Zhang, S.~Agarwal, K.~Slama, A.~Gray, et al.
Training language models to follow instructions with human feedback.
In \textit{Advances in Neural Information Processing Systems}, 2022.

\bibitem{sdxl}
D.~Podell, Z.~English, K.~Lacey, A.~Blattmann, T.~Dockhorn, J.~M\"uller, J.~Penna, and R.~Rombach.
SDXL: improving latent diffusion models for high-resolution image synthesis.
In \textit{Proceedings of the International Conference on Learning Representations}, 2024.

\bibitem{visual-chatgpt}
C.~Wu, S.~Yin, W.~Qi, X.~Wang, Z.~Tang, and N.~Duan.
Visual ChatGPT: talking, drawing and editing with visual foundation models.
\textit{arXiv preprint arXiv:2303.04671}, 2023.

\bibitem{pickapic}
Y.~Kirstain, A.~Polyak, U.~Singer, S.~Matiana, J.~Penna, and O.~Levy.
Pick-a-Pic: an open dataset of user preferences for text-to-image generation.
In \textit{Advances in Neural Information Processing Systems}, 2023.

\bibitem{recaption}
E.~Segalis, D.~Valevski, D.~Lumen, Y.~Matias, and Y.~Leviathan.
A picture is worth a thousand words: principled recaptioning improves image generation.
\textit{arXiv preprint arXiv:2310.16656}, 2023.

\bibitem{rewardbp}
M.~Prabhudesai, A.~Goyal, D.~Pathak, and K.~Fragkiadaki.
Aligning text-to-image diffusion models with reward backpropagation.
\textit{arXiv preprint arXiv:2310.03739}, 2023.

\bibitem{ddpm}
J.~Ho, A.~Jain, and P.~Abbeel.
Denoising diffusion probabilistic models.
In \textit{Advances in Neural Information Processing Systems}, 2020.

\bibitem{flux}
Black Forest Labs, S.~Batifol, A.~Blattmann, F.~Boesel, S.~Consul, C.~Diagne, T.~Dockhorn, J.~English, Z.~English, P.~Esser, S.~Kulal, K.~Lacey, Y.~Levi, C.~Li, D.~Lorenz, J.~M\"{u}ller, D.~Podell, R.~Rombach, H.~Saini, A.~Sauer, and L.~Smith.
FLUX.1 Kontext: flow matching for in-context image generation and editing in latent space.
\textit{arXiv preprint arXiv:2506.15742}, 2025.

\bibitem{imagereward}
J.~Xu, X.~Liu, Y.~Wu, Y.~Tong, Q.~Li, M.~Ding, J.~Tang, and Y.~Dong.
ImageReward: learning and evaluating human preferences for text-to-image generation.
\textit{arXiv preprint arXiv:2304.05977}, 2023.

\bibitem{dalle}
A.~Ramesh, M.~Pavlov, G.~Goh, S.~Gray, C.~Voss, A.~Radford, M.~Chen, and I.~Sutskever.
Zero-shot text-to-image generation.
In \textit{Proceedings of the International Conference on Machine Learning}, 2021.

\bibitem{ediffi}
Y.~Balaji, S.~Nah, X.~Huang, A.~Vahdat, J.~Song, Q.~Zhang, K.~Kreis, M.~Aittala, T.~Aila, S.~Laine, B.~Catanzaro, T.~Karras, and M.-Y.~Liu.
eDiff-I: text-to-image diffusion models with an ensemble of expert denoisers.
\textit{arXiv preprint arXiv:2211.01324}, 2022.

\bibitem{ldm}
R.~Rombach, A.~Blattmann, D.~Lorenz, P.~Esser, and B.~Ommer.
High-resolution image synthesis with latent diffusion models.
In \textit{Proceedings of the IEEE/CVF Conference on Computer Vision and Pattern Recognition}, 2022.

\bibitem{aligninghf}
K.~Lee, H.~Liu, M.~Ryu, O.~Watkins, Y.~Du, C.~Boutilier, P.~Abbeel, M.~Ghavamzadeh, and S.~S.~Gu.
Aligning text-to-image models using human feedback.
\textit{arXiv preprint arXiv:2302.12192}, 2023.

\bibitem{diffusiondpo}
B.~Wallace, M.~Dang, R.~Rafailov, L.~Zhou, A.~Lou, S.~Purushwalkam, S.~Ermon, C.~Xiong, S.~Joty, and N.~Naik.
Diffusion model alignment using direct preference optimization.
In \textit{Proceedings of the IEEE/CVF Conference on Computer Vision and Pattern Recognition}, 2024.

\bibitem{glide}
A.~Nichol, P.~Dhariwal, A.~Ramesh, P.~Shyam, P.~Mishkin, B.~McGrew, I.~Sutskever, and M.~Chen.
GLIDE: towards photorealistic image generation and editing with text-guided diffusion models.
In \textit{Proceedings of the International Conference on Machine Learning}, 2022.

\bibitem{freedom}
J.~Yu, Y.~Wang, C.~Zhao, B.~Ghanem, and J.~Zhang.
FreeDoM: training-free energy-guided conditional diffusion model.
In \textit{Proceedings of the IEEE/CVF International Conference on Computer Vision}, 2023.

\bibitem{spo}
Z.~Liang, X.~Han, X.~Wang, G.~Song, Y.~Liu, and H.~Li.
Aesthetic post-training diffusion models from generic preferences with step-by-step preference optimization.
In \textit{Proceedings of the IEEE/CVF Conference on Computer Vision and Pattern Recognition}, 2025.

\bibitem{rl-diffusion}
K.~Black, M.~Janner, Y.~Du, I.~Kostrikov, and S.~Levine.
Training diffusion models with reinforcement learning.
In \textit{Proceedings of the International Conference on Learning Representations}, 2024.

\bibitem{imagen}
C.~Saharia, W.~Chan, S.~Saxena, L.~Li, J.~Whang, E.~Denton, S.~Ghasemipour, B.~Karagol Ayan, S.~Mahdavi, R.~G. Lopes, T.~Salimans, J.~Ho, D.~J. Fleet, and M.~Norouzi.
Photorealistic text-to-image diffusion models with deep language understanding.
In \textit{Advances in Neural Information Processing Systems}, 2022.

\bibitem{prompts-opt}
Y.~Hao, Z.~Chi, L.~Dong, and F.~Wei.
Optimizing prompts for text-to-image generation.
In \textit{Advances in Neural Information Processing Systems}, 2023.

\bibitem{d3po}
K.~Yang, J.~Tao, J.~Lyu, C.~Ge, J.~Chen, Q.~Li, W.~Shen, X.~Zhu, and X.~Li.
Using human feedback to fine-tune diffusion models without any reward model.
In \textit{Proceedings of the IEEE/CVF Conference on Computer Vision and Pattern Recognition}, 2024.

\bibitem{hps}
X.~Wu, K.~Sun, F.~Zhu, R.~Zhao, and H.~Li.
Human preference score: better aligning text-to-image models with human preference.
In \textit{Proceedings of the IEEE/CVF International Conference on Computer Vision}, 2023.

\bibitem{hpsv2}
X.~Wu, Y.~Hao, K.~Sun, Y.~Chen, F.~Zhu, R.~Zhao, and H.~Li.
Human preference score v2: a solid benchmark for evaluating human preferences of text-to-image synthesis.
\textit{arXiv preprint arXiv:2306.09341}, 2023.

\bibitem{draft}
K.~Clark, P.~Vicol, K.~Swersky, and D.~J.~Fleet.
Directly fine-tuning diffusion models on differentiable rewards.
\textit{arXiv preprint arXiv:2309.17400}, 2023.

\bibitem{dno}
Z.~Tang, J.~Peng, J.~Tang, M.~Hong, F.~Wang, and T.-H.~Chang.
Tuning-free alignment of diffusion models with direct noise optimization.
\textit{arXiv preprint arXiv:2405.18881}, 2024.

\bibitem{deckers2023promptopt}
N.~Deckers, J.~Peters, and M.~Potthast.
Manipulating embeddings of Stable Diffusion prompts.
\textit{arXiv preprint arXiv:2308.12059}, 2023.

\bibitem{diffkto}
S.~Li, K.~Kallidromitis, A.~Gokul, Y.~Kato, and K.~Kozuka.
Aligning diffusion models by optimizing human utility.
\textit{arXiv preprint arXiv:2404.04465}, 2024.

\bibitem{ddim}
J.~Song, C.~Meng, and S.~Ermon.
Denoising diffusion implicit models.
In \textit{Proceedings of the International Conference on Learning Representations}, 2021.



\end{thebibliography}

}
\end{document}